\documentclass[10pt,journal,compsoc, onecolumn]{IEEEtran}
\usepackage{cite}
\usepackage[pdftex]{graphicx}
\DeclareGraphicsExtensions{.pdf,.jpeg,.png}
\usepackage{array}
\usepackage{amsmath}
\usepackage{amssymb}
\usepackage[table]{xcolor}       %
\usepackage{colortbl}      %
\usepackage{mdwmath}
\usepackage{mdwtab}
\usepackage{eqparbox}
\usepackage{url}
\usepackage{hyperref}
\usepackage{multirow}
\usepackage{algorithm}
\usepackage{algpseudocode}
\usepackage{lineno} 
\usepackage{booktabs}
\usepackage{rotating} 

\usepackage[most]{tcolorbox}
\usepackage{enumitem}

\definecolor{promptFrame}{HTML}{999999}
\definecolor{promptBody}{HTML}{f2f2f2}

\definecolor{cellHigh}{HTML}{2ca02c}    
\definecolor{cellMid}{HTML}{ffdd57}     
\definecolor{cellLow}{HTML}{d62728}     
\newcommand{\acc}[1]{%
  \pgfmathtruncatemacro{\acclevel}{#1 < 60 ? 1 : (#1 < 75 ? 2 : (#1 < 85 ? 3 : 4))}%
  \ifnum\acclevel=1 \cellcolor{cellLow!40}\fi
  \ifnum\acclevel=2 \cellcolor{cellMid!40}\fi
  \ifnum\acclevel=3 \cellcolor{cellMid!20}\fi
  \ifnum\acclevel=4 \cellcolor{cellHigh!25}\fi
  #1}
\newcommand{\best}[1]{\textbf{#1}}

\begin{document}

\title{Benchmarking and Adapting On-Device LLMs for Clinical Decision Support}

\author{Alif Munim$^*$, Omar Ibrahim$^*$, Alhusain Abdalla$^*$, Jun Ma$^*$, Meng Wei, Shuolin Yin, Leo Chen, and Bo Wang 

\IEEEcompsocitemizethanks{
\IEEEcompsocthanksitem Alif Munim is with AI Collaborative Centre, University Health Network, Toronto, Canada. ($*$ Equal Contribution)
\IEEEcompsocthanksitem Omar Ibrahim is with AI Collaborative Centre, University Health Network, Toronto, Canada. ($*$ Equal Contribution)
\IEEEcompsocthanksitem Alhusain Abdalla is with AI Collaborative Centre, University Health Network, Toronto, Canada. ($*$ Equal Contribution)
\IEEEcompsocthanksitem Jun Ma is with AI Collaborative Centre and Princess Margaret Cancer Centre, University Health Network, Toronto, Canada. ($*$ Equal Contribution)
\IEEEcompsocthanksitem Meng Wei is with AI Collaborative Centre and Princess Margaret Cancer Centre, University Health Network, Toronto, Canada.
\IEEEcompsocthanksitem Shuolin Yin is with Department of Electrical and Computer Engineering, University of Toronto, Toronto, Canada.
\IEEEcompsocthanksitem Leo Chen is with Division of Urology, Department of Surgery, St. Michael’s Hospital, Unity Health Toronto and University of Toronto, Toronto, Canada
\IEEEcompsocthanksitem Bo Wang (Corresponding Author) is with Peter Munk Cardiac Centre, University Health Network; Department of Laboratory Medicine and Pathobiology and Department of Computer Science, University of Toronto;  Vector Institute, Toronto, Canada. 
E-mail: bowang@vectorinstitute.ai}
}

\IEEEtitleabstractindextext{%
\begin{abstract}
Large language models (LLMs) have rapidly advanced in clinical decision-making, yet the deployment of proprietary systems is hindered by privacy concerns and reliance on cloud-based infrastructure. Open-source alternatives allow local inference but often have large model sizes that limit their use in resource-constrained clinical settings. Here, we benchmark on-device LLMs from the gpt-oss (20b, 120b), Qwen3.5 (9B, 27B, 35B), and Gemma 4 (31B) families across three representative clinical tasks: general disease diagnosis, specialty-specific (ophthalmology) diagnosis and management, and simulation of human expert grading and evaluation. We compare their performance with state-of-the-art proprietary models (GPT-5.1, GPT-5-mini, and Gemini 3.1 Pro) and a leading open-source model (DeepSeek-R1), and we further evaluate the adaptability of on-device systems by fine-tuning gpt-oss-20b and Qwen3.5-35B on general diagnostic data. Across tasks, on-device models achieve performance comparable to or exceeding DeepSeek-R1 and GPT-5-mini despite being substantially smaller. In addition, fine-tuning remarkably improves diagnostic accuracy, with the fine-tuned Qwen3.5-35B reaching 87.9\% and approaching the proprietary GPT-5.1 (89.4\%). Among base on-device models, Gemma 4 31B achieved the strongest general diagnostic accuracy at 86.5\%, exceeding GPT-5-mini and approaching the fine-tuned Qwen3.5-35B. Error characterization revealed that 87.2\% of diagnostic errors across all models were clinically plausible differentials rather than off-topic predictions, and upper-bound analysis showed up to 93.2\% attainable accuracy through improved answer selection. These findings highlight the potential of on-device LLMs to deliver accurate, adaptable, and privacy-preserving clinical decision support, offering a practical pathway for broader integration of LLMs into routine clinical practice.
\end{abstract}

}

\maketitle

\IEEEdisplaynontitleabstractindextext


\IEEEpeerreviewmaketitle

\section*{Introduction}
Large language models (LLMs) are rapidly transforming the landscape of clinical medicine~\cite{thirunavukarasu2023llmedicine, omiye2024pitfalls}. Trained on massive corpora of general-domain and biomedical text, these models have demonstrated emergent reasoning abilities that enable comprehensive summaries from medical dialogue~\cite{NMed-LLMSummary}, disease diagnosis, and treatment planning~\cite{MedFound}. For example, Med-PaLM~\cite{MedPALM1,MedPALM2} and AMIE~\cite{AMIE-diagnosis,AMIE-conversation} achieved near-clinician performance in open-ended medical question answering, differential diagnosis, and patient consultation simulations~\cite{qiu2025medrench}. In real-world evaluations~\cite{openai-AIConsult, goh2025gpt4rct}, LLM-assisted physicians exhibited improved diagnostic accuracy and management decisions across diverse clinical scenarios. Together, these advances illustrate the growing potential of LLMs to augment clinical expertise and support decision-making at the point of care.

Despite these achievements, the translation of LLMs into real-world clinical workflows remains limited~\cite{hager2024limitations}. Most frontier LLMs are proprietary, cloud-hosted, and trained on non-transparent datasets, creating challenges related to data privacy, regulatory compliance, reproducibility, and cost. The transmission of patient information to external servers conflicts with data-governance policies in many health institutions~\cite{kim2025privacy, clusmann2025implementing}. Moreover, the high cost associated with developing, training, and deploying large-scale proprietary models can be prohibitive for many healthcare institutions. 

The open-source community has developed increasingly capable models that approach the performance of proprietary systems while allowing full local control. Models such as DeepSeek-R1~\cite{DeepSeek-R1} have demonstrated competitive reasoning and clinical comprehension across diagnostic and treatment-related tasks~\cite{eval-deepseek,eval-deepseek125Patients, wang2025jarvis, moell2025deepseek, deepseek2025cme}. 
However, the adoption of DeepSeek-R1 in clinical environments is constrained because of the large model size of 671 billion parameters, which requires substantial computational costs. The models of this scale not only hinder deployment in clinics with limited access to computing resources, but also make fine-tuning for adaptation in evolving clinical domains prohibitively expensive. 
These limitations underscore the need for smaller, efficient models that retain clinical reasoning ability while remaining feasible for on-premise use~\cite{garg2025rise, builtjes2025leveraging}.

\begin{figure}[htbp]
\centering
\includegraphics[width=\textwidth]{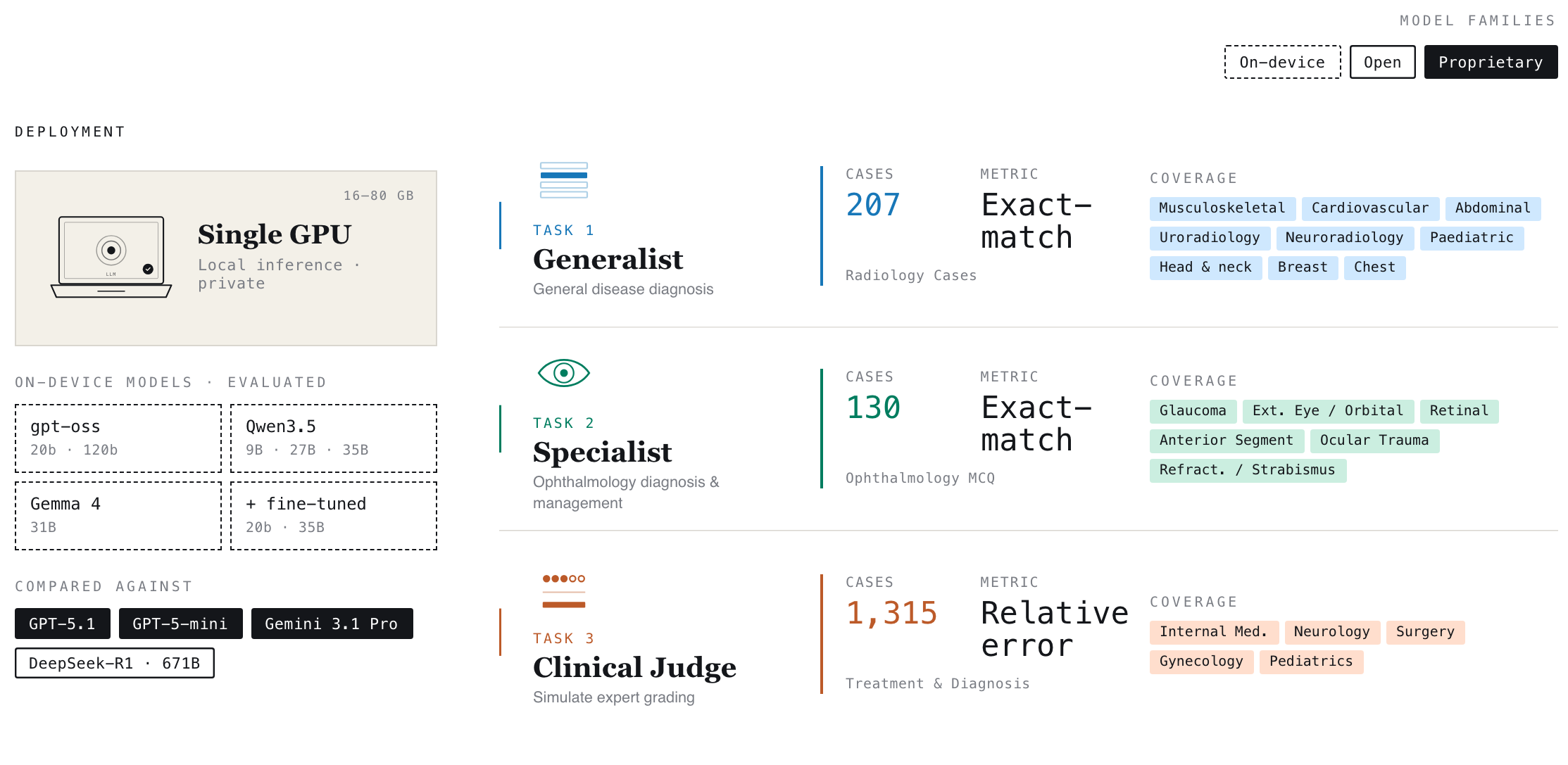} 
\caption{\textbf{Overview of the benchmark framework.} This study compares on-device LLMs (gpt-oss, Qwen3.5, and Gemma 4) with state-of-the-art open-source (DeepSeek-R1) and proprietary LLMs (GPT-5.1, GPT-5-mini, Gemini 3.1 Pro) across general disease diagnosis, specialty diagnosis and management (ophthalmology), and clinical judgment simulation.}
\label{fig:1}
\end{figure}

The recent advancements in on-device LLMs offer a promising solution to the challenges posed by both large proprietary and open-source models~\cite{zhu2025empowering, evangelista2026graphrag}. In particular, the gpt-oss family~\cite{gpt-oss} represents this new generation of efficient and privacy-preserving architectures. The gpt-oss-20b model and its larger counterpart, gpt-oss-120b, are designed and optimized for deployment on a single consumer GPU with 16GB and 80GB memory, respectively. Similarly, the Qwen3.5 family of open-weight models (9B, 27B, and 35B parameters) offers an alternative on-device architecture suitable for resource-constrained deployment.
This study aims to determine whether on-device LLMs can be practically used for clinical decision support. We systematically benchmark gpt-oss (20b, 120b), Qwen3.5 (9B, 27B, 35B), and Gemma 4 (31B) models across three representative tasks: general disease diagnosis, (ophthalmology) specialty-specific disease diagnosis and management, and simulation of human clinician judgement and evaluation (Fig.~\ref{fig:1}). To establish a clear performance landscape, we compare these on-device models against leading proprietary models (GPT-5.1, GPT-5-mini, and Gemini 3.1 Pro) and a strong open-source model (DeepSeek-R1). Beyond aggregate accuracy, we characterize the nature of residual errors through a structured taxonomy and quantify the recoverable performance ceiling via upper-bound analysis across independent runs.

\section*{Results}
\subsection*{Dataset and evaluation methods} 
We mainly focus on assessing the performance of LLMs on disease diagnosis, treatment recommendations, and simulating expert judgment for open-ended questions, as these are common tasks in clinical practice. We curated three datasets and benchmarked model performance on three scenarios: LLM-as-a-generalist, LLM-as-a-specialist, and LLM-as-a-clinical-judge (Fig.~\ref{fig:1}).  

Specifically, to assess the capability of LLMs for general disease diagnosis, we collected 207 case reports from the Eurorad library~\cite{kottlors2025eurorad} where each case contains clinical history, findings from associated medical images (e.g., computed tomography (CT), magnetic resonance imaging  (MRI), and ultrasound), and a fine-grained differential diagnosis list (Methods). The task is to select the diagnosis from the list based on patient history and imaging findings (Supplementary Prompt 1). 

The second dataset was from an ophthalmology multiple-choice question dataset~\cite{ophthalmologyQA}, aiming to evaluate the specialty-specific performance of LLMs (Methods). The dataset contains 39 diagnosis questions and 91 management questions. Each question contains patient sex, age, and examinations findings, such as visual acuity, intraocular pressure, and fundus examinations, and a list of five to nine answer options. The task is to select correct answers, and each question may have multiple correct answers (Supplementary Prompt 2).  

The third dataset was adapted from the existing benchmarking study of the DeepSeek-R1 model~\cite{eval-deepseek125Patients}, including 1315 cases with human expert scores from 125 patients across five specialties (internal medicine, neurology, surgery, gynecology, and pediatrics) (Methods). LLMs are tasked to simulate human experts to assess given diagnoses from existing LLMs and assign a score from 1 to 5 based on a predefined rubric. We compare these LLM-generated scores against ground-truth human expert scores  (Supplementary Prompt 3-4). 

 For the LLM-as-a-generalist and LLM-as-a-specialist tasks, the evaluation was based on exact match accuracy. For the LLM-as-a-clinical-judge task, we computed the relative error between the clinicians' scores and the LLM's scores. 
To ensure rigorous zero-shot evaluation and prevent data leakage, all three benchmark datasets were curated from cases released after the training data cutoff for all evaluated LLMs.

\begin{figure}[!htbp]
\centering
\enlargethispage{2cm}
\vspace*{-1.5cm}
\includegraphics[width=\textwidth,height=0.92\textheight,keepaspectratio]{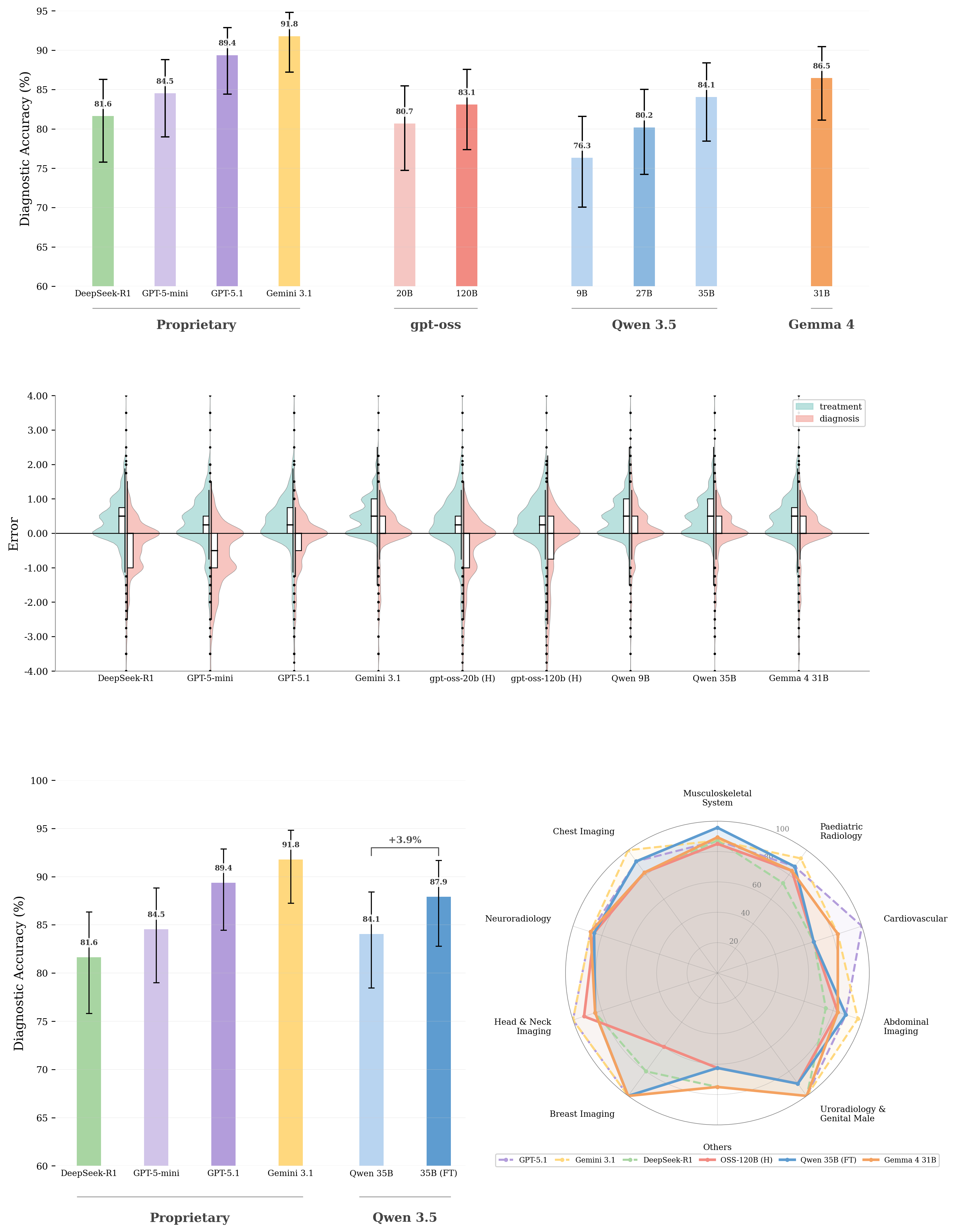}
\caption{\textbf{Zero-shot and fine-tuning performance of on-device LLMs.}
\textbf{a,} Results of LLM-as-a-generalist: diagnosis accuracy on a wide range of radiological cases (N=207).
\textbf{b,} Results of LLM-as-a-clinical-judge: violin plots comparing the relative error for disease diagnosis and treatment open-ended question assessment (N=1315).
\textbf{c,} Fine-tuned Qwen3.5-35B model accuracy compared to proprietary LLMs on the disease differential diagnosis task.
\textbf{d,} Model performance across 10 radiological sub-specialties.
}\label{fig:2}
\end{figure}

\begin{table}[!htbp]
\centering
\scriptsize
\setlength{\tabcolsep}{3pt}
\caption{\textbf{LLM-as-a-Generalist Task: Comparative diagnostic accuracy (\%) across radiological anatomical subgroups.}
Performance is evaluated using self-consistency majority voting. The best result per row is \textbf{bolded}. For gpt-oss models, the best-performing reasoning effort variant is shown; 20b-FT denotes the fine-tuned gpt-oss-20b (Medium effort). Full 95\% CIs are provided in Supplementary Table~1.}
\label{tab:generalist_results}
\begin{tabular}{l ccc @{\hskip 6pt} c @{\hskip 6pt} ccc @{\hskip 6pt} cccc @{\hskip 6pt} c}
\toprule
& \multicolumn{3}{c}{\textbf{Proprietary}} & \textbf{Open} & \multicolumn{3}{c}{\textbf{gpt-oss}} & \multicolumn{4}{c}{\textbf{Qwen3.5}} & \textbf{Gemma 4} \\
\cmidrule(lr){2-4} \cmidrule(lr){5-5} \cmidrule(lr){6-8} \cmidrule(lr){9-12} \cmidrule(lr){13-13}
\textbf{Category} &
\textbf{GPT-5.1} & \textbf{GPT-5-mini} & \textbf{Gemini-3.1} & \textbf{DS-R1} &
\textbf{20b} & \textbf{120b} & \textbf{20b-FT} &
\textbf{9B} & \textbf{27B} & \textbf{35B} & \textbf{35B-FT} & \textbf{31B} \\
\midrule
Musculoskeletal   & \acc{87.2} & \acc{83.0} & \acc{87.2} & \acc{87.2} & \acc{87.2} & \acc{85.1} & \acc{91.5} & \acc{89.4} & \acc{87.2} & \acc{89.4} & \best{\acc{95.7}} & \acc{89.4} \\
Cardiovascular    & \best{\acc{100.0}} & \acc{83.3} & \acc{83.3} & \acc{66.7} & \acc{50.0} & \acc{66.7} & \acc{83.3} & \acc{66.7} & \acc{50.0} & \acc{50.0} & \acc{66.7} & \acc{83.3} \\
Abdominal         & \acc{88.9} & \acc{88.9} & \best{\acc{97.2}} & \acc{75.0} & \acc{80.6} & \acc{83.3} & \acc{86.1} & \acc{72.2} & \acc{83.3} & \acc{83.3} & \acc{88.9} & \acc{83.3} \\
Uroradiology      & \best{\acc{100.0}} & \acc{90.0} & \best{\acc{100.0}} & \best{\acc{100.0}} & \best{\acc{100.0}} & \acc{90.0} & \best{\acc{100.0}} & \best{\acc{100.0}} & \acc{90.0} & \best{\acc{100.0}} & \acc{90.0} & \best{\acc{100.0}} \\
Neuroradiology    & \best{\acc{87.8}} & \acc{82.9} & \best{\acc{87.8}} & \acc{85.4} & \acc{85.4} & \acc{85.4} & \best{\acc{87.8}} & \acc{70.7} & \acc{78.0} & \acc{82.9} & \acc{85.4} & \best{\acc{87.8}} \\
Paediatric        & \acc{86.7} & \acc{80.0} & \best{\acc{93.3}} & \acc{73.3} & \acc{70.0} & \acc{83.3} & \acc{73.3} & \acc{63.3} & \acc{70.0} & \acc{80.0} & \acc{86.7} & \acc{83.3} \\
Head \& neck      & \best{\acc{100.0}} & \best{\acc{100.0}} & \best{\acc{100.0}} & \acc{84.6} & \acc{92.3} & \acc{92.3} & \acc{92.3} & \acc{84.6} & \acc{92.3} & \acc{92.3} & \acc{84.6} & \acc{84.6} \\
Breast            & \best{\acc{100.0}} & \acc{80.0} & \best{\acc{100.0}} & \acc{80.0} & \acc{40.0} & \acc{60.0} & \acc{80.0} & \acc{80.0} & \best{\acc{100.0}} & \best{\acc{100.0}} & \best{\acc{100.0}} & \best{\acc{100.0}} \\
Chest             & \acc{90.9} & \acc{81.8} & \best{\acc{100.0}} & \acc{81.8} & \acc{81.8} & \acc{81.8} & \acc{90.9} & \acc{63.6} & \acc{63.6} & \acc{72.7} & \acc{90.9} & \acc{81.8} \\
Others            & \best{\acc{75.0}} & \best{\acc{75.0}} & \best{\acc{75.0}} & \best{\acc{75.0}} & \acc{62.5} & \acc{62.5} & \best{\acc{75.0}} & \best{\acc{75.0}} & \best{\acc{75.0}} & \best{\acc{75.0}} & \acc{62.5} & \best{\acc{75.0}} \\
\midrule
\textbf{Average}  & \acc{89.4} & \acc{84.5} & \best{\acc{91.8}} & \acc{81.6} & \acc{80.7} & \acc{83.1} & \acc{86.5} & \acc{76.3} & \acc{80.2} & \acc{84.1} & \acc{87.9} & \acc{86.5} \\
\bottomrule
\end{tabular}
\end{table}

\subsection*{On-device LLMs show competitive zero-shot performance}
Fig.~\ref{fig:2}a and Table~\ref{tab:generalist_results} compare the diagnostic accuracy of the evaluated models for general disease diagnosis across radiological case reports. Among the proprietary frontier models, Gemini 3.1 Pro achieved the highest overall accuracy of 91.8\% (95\% CI: 87.2--94.8\%), followed by GPT-5.1 at 89.4\% (95\% CI: 84.4--92.9\%). The on-device models demonstrated strong competitiveness relative to these baselines. Specifically, the gpt-oss-120b model achieved an accuracy of 83.1\% (95\% CI: 77.4--87.6\%), while the highly efficient gpt-oss-20b model secured 80.7\% (95\% CI: 74.8--85.5\%), matching the performance of the significantly larger DeepSeek-R1 (81.6\%; 95\% CI: 75.8--86.3\%; $p>0.05$). Among the Qwen3.5 on-device models, the 35B variant achieved the highest base accuracy at 84.1\% (95\% CI: 78.5--88.4\%), comparable to GPT-5-mini (84.5\%; 95\% CI: 79.0--88.8\%; $p>0.05$) and exceeding DeepSeek-R1. The Qwen3.5 27B model achieved 80.2\% (95\% CI: 74.2--85.1\%), comparable to both gpt-oss-20b and DeepSeek-R1 ($p>0.05$), while the smallest 9B variant reached 76.3\% (95\% CI: 70.1--81.6\%).

Gemma 4 31B achieved the strongest base on-device accuracy at 86.5\% (95\% CI: 81.1--90.5\%), exceeding GPT-5-mini (84.5\%; $p>0.05$) and surpassing the strongest Qwen3.5 base variant (35B, 84.1\%) by 2.4 percentage points. This result places Gemma 4 31B between GPT-5-mini and GPT-5.1 (89.4\%) on the general diagnosis task, and significantly above Qwen3.5 9B ($p<0.05$). Subspecialty analysis (Table~\ref{tab:generalist_results}) shows uniformly strong performance, with perfect scores on uroradiology and breast imaging.

\begin{table}[!htbp]
\centering
\scriptsize
\setlength{\tabcolsep}{3.5pt}
\caption{\textbf{LLM-as-a-Specialist Task: Comparative accuracy (\%) on ophthalmology diagnosis and management tasks.} The best result per row is \textbf{bolded}. For gpt-oss models, the best-performing reasoning effort variant is shown (20b=High, 120b=Medium). Full 95\% CIs are provided in Supplementary Table~2; reasoning effort analysis in Supplementary Table~7.}
\label{tab:ophthalmology_results}
\begin{tabular}{ll ccc @{\hskip 6pt} c @{\hskip 6pt} cc @{\hskip 6pt} ccc @{\hskip 6pt} c}
\toprule
& & \multicolumn{3}{c}{\textbf{Proprietary}} & \textbf{Open} & \multicolumn{2}{c}{\textbf{gpt-oss}} & \multicolumn{3}{c}{\textbf{Qwen3.5}} & \textbf{Gemma 4} \\
\cmidrule(lr){3-5} \cmidrule(lr){6-6} \cmidrule(lr){7-8} \cmidrule(lr){9-11} \cmidrule(lr){12-12}
\textbf{Type} & \textbf{Topic} & \textbf{GPT-5.1} & \textbf{GPT-5-mini} & \textbf{Gemini-3.1} & \textbf{DS-R1} & \textbf{20b} & \textbf{120b} & \textbf{9B} & \textbf{27B} & \textbf{35B} & \textbf{31B} \\
\midrule
\multirow{6}{*}{\rotatebox{90}{\textbf{Diagnosis}}}
& Glaucoma            & \best{\acc{100.0}} & \best{\acc{100.0}} & \best{\acc{100.0}} & \best{\acc{100.0}} & \acc{85.7} & \acc{85.7} & \best{\acc{100.0}} & \best{\acc{100.0}} & \best{\acc{100.0}} & \acc{85.7} \\
& Ext.\ Eye/Orbital  & \best{\acc{90.0}} & \acc{80.0} & \best{\acc{90.0}} & \best{\acc{90.0}} & \acc{80.0} & \acc{80.0} & \acc{70.0} & \acc{70.0} & \acc{70.0} & \acc{80.0} \\
& Retinal            & \best{\acc{100.0}} & \best{\acc{100.0}} & \best{\acc{100.0}} & \best{\acc{100.0}} & \acc{66.7} & \acc{66.7} & \acc{66.7} & \best{\acc{100.0}} & \best{\acc{100.0}} & \acc{66.7} \\
& Anterior Segment   & \acc{50.0} & \best{\acc{75.0}} & \best{\acc{75.0}} & \best{\acc{75.0}} & \acc{62.5} & \best{\acc{75.0}} & \best{\acc{75.0}} & \best{\acc{87.5}} & \best{\acc{75.0}} & \best{\acc{75.0}} \\
& Ocular Trauma      & \best{\acc{100.0}} & \acc{83.3} & \best{\acc{100.0}} & \best{\acc{100.0}} & \acc{83.3} & \best{\acc{100.0}} & \acc{83.3} & \best{\acc{100.0}} & \best{\acc{100.0}} & \best{\acc{100.0}} \\
& Refract./Strabismus & \best{\acc{100.0}} & \best{\acc{100.0}} & \best{\acc{100.0}} & \best{\acc{100.0}} & \best{\acc{100.0}} & \best{\acc{100.0}} & \best{\acc{100.0}} & \best{\acc{100.0}} & \best{\acc{100.0}} & \best{\acc{100.0}} \\
\cmidrule{2-12}
& \textit{Average}   & \textit{\acc{87.2}} & \textit{\acc{87.2}} & \textit{\best{\acc{92.3}}} & \textit{\best{\acc{92.3}}} & \textit{\acc{79.5}} & \textit{\acc{84.6}} & \textit{\acc{82.1}} & \textit{\best{\acc{89.7}}} & \textit{\acc{87.2}} & \textit{\acc{84.6}} \\
\midrule
\multirow{6}{*}{\rotatebox{90}{\textbf{Management}}}
& Glaucoma            & \best{\acc{85.7}} & \acc{64.3} & \best{\acc{85.7}} & \acc{71.4} & \acc{50.0} & \acc{78.6} & \acc{64.3} & \best{\acc{85.7}} & \best{\acc{85.7}} & \acc{71.4} \\
& Ext.\ Eye/Orbital  & \acc{71.4} & \acc{57.1} & \acc{64.3} & \acc{78.6} & \acc{64.3} & \acc{71.4} & \acc{78.6} & \acc{71.4} & \best{\acc{92.9}} & \acc{78.6} \\
& Retinal            & \acc{37.5} & \acc{25.0} & \acc{37.5} & \acc{37.5} & \acc{37.5} & \acc{37.5} & \acc{37.5} & \acc{25.0} & \acc{37.5} & \acc{37.5} \\
& Anterior Segment   & \best{\acc{82.4}} & \acc{64.7} & \best{\acc{82.4}} & \acc{70.6} & \acc{76.5} & \acc{76.5} & \acc{58.8} & \acc{64.7} & \acc{64.7} & \acc{58.8} \\
& Ocular Trauma      & \acc{65.4} & \acc{57.7} & \acc{65.4} & \best{\acc{76.9}} & \acc{50.0} & \best{\acc{76.9}} & \acc{57.7} & \acc{61.5} & \acc{57.7} & \acc{69.2} \\
& Refract./Strabismus & \acc{83.3} & \acc{75.0} & \best{\acc{91.7}} & \best{\acc{91.7}} & \acc{83.3} & \best{\acc{91.7}} & \best{\acc{91.7}} & \best{\acc{91.7}} & \acc{83.3} & \best{\acc{91.7}} \\
\cmidrule{2-12}
& \textit{Average}   & \textit{\acc{72.5}} & \textit{\acc{59.3}} & \textit{\acc{72.5}} & \textit{\best{\acc{73.6}}} & \textit{\acc{60.4}} & \textit{\best{\acc{74.7}}} & \textit{\acc{64.8}} & \textit{\acc{68.1}} & \textit{\acc{70.3}} & \textit{\acc{69.2}} \\
\midrule
\multicolumn{2}{l}{\textbf{Overall}} & \acc{76.9} & \acc{67.7} & \acc{78.5} & \best{\acc{79.2}} & \acc{66.2} & \acc{77.7} & \acc{70.0} & \acc{74.6} & \acc{75.4} & \acc{73.8} \\
\bottomrule
\end{tabular}
\end{table}

Table~\ref{tab:ophthalmology_results} presents the model performance of the LLM-as-a-specialist task on the ophthalmology QA dataset. The open-source model DeepSeek-R1 achieved the highest overall accuracy of 79.2\% (95\% CI: 71.5--85.3\%), demonstrating particularly strong capabilities in diagnosis (92.3\%; 95\% CI: 79.7--97.3\%) and management (73.6\%; 95\% CI: 63.7--81.6\%). Remarkably, the on-device gpt-oss-120b model achieved a rivaling performance with an overall accuracy of 77.7\% (95\% CI: 69.8--84.0\%), surpassing the proprietary GPT-5.1 (76.9\%) and GPT-5-mini (67.7\%) baselines. Gemma 4 31B achieved 73.8\% overall (84.6\% diagnosis, 69.2\% management), comparable to gpt-oss-120b (70.0\%) and GPT-5.1 (76.9\%). This performance underscores the capacity of locally deployed large models to rival state-of-the-art proprietary systems in specialty-specific medical reasoning. While the smaller gpt-oss-20b showed a performance gap at 66.2\% (95\% CI: 57.7--73.7\%), it remained competitive in specific subtasks (e.g., Refractive Disorders or Strabismus). Across all evaluated architectures, models consistently demonstrated higher performance in diagnosis compared to patient management, reflecting the increased complexity of treatment planning in specialty care.

\begin{table}[!htbp]
\centering
\scriptsize
\setlength{\tabcolsep}{3.5pt}
\caption{\textbf{LLM-as-a-Clinical-Judge Task: Median error relative to human expert scores across five specialties.} A value of 0.00 indicates perfect alignment; negative values indicate underestimation. The value closest to 0.00 per row is \textbf{bolded}. For gpt-oss models, the best-performing reasoning effort variant is shown (Low for both). Full IQR values are provided in Supplementary Table~3; reasoning effort analysis in Supplementary Table~8.}
\label{tab:specialty_results_with_total}
\begin{tabular}{ll ccc @{\hskip 6pt} c @{\hskip 6pt} cc @{\hskip 6pt} ccc @{\hskip 6pt} c}
\toprule
& & \multicolumn{3}{c}{\textbf{Proprietary}} & \textbf{Open} & \multicolumn{2}{c}{\textbf{gpt-oss}} & \multicolumn{3}{c}{\textbf{Qwen3.5}} & \textbf{Gemma 4} \\
\cmidrule(lr){3-5} \cmidrule(lr){6-6} \cmidrule(lr){7-8} \cmidrule(lr){9-11} \cmidrule(lr){12-12}
\textbf{Type} & \textbf{Specialty} & \textbf{GPT-5.1} & \textbf{Gemini-3.1} & \textbf{GPT-5-mini} & \textbf{DS-R1} & \textbf{20b} & \textbf{120b} & \textbf{9B} & \textbf{27B} & \textbf{35B} & \textbf{31B} \\
\midrule
\multirow{6}{*}{\rotatebox{90}{\textbf{Treatment}}}
& Gynecology & 0.25 & 0.50 & 0.25 & 0.38 & 0.17 & \best{0.00} & 0.50 & 0.50 & 0.50 & 0.25 \\
& Internal Med. & 0.33 & 0.50 & \best{0.00} & 0.17 & \best{0.00} & \best{0.00} & 0.33 & 0.42 & 0.50 & 0.42 \\
& Neurology & 0.50 & 0.50 & 0.33 & 0.21 & \best{0.00} & 0.17 & 0.50 & 0.50 & 0.50 & 0.50 \\
& Pediatrics & 0.25 & 0.50 & 0.21 & 0.17 & \best{0.00} & 0.17 & 0.33 & 0.33 & 0.50 & 0.50 \\
& Surgery & 0.50 & 0.50 & 0.50 & 0.50 & \best{0.25} & 0.33 & 0.50 & 0.50 & 0.50 & 0.50 \\
\cmidrule{2-12}
& \textit{Overall} & \textit{0.33} & \textit{0.50} & \textit{0.25} & \textit{0.29} & \textit{\best{0.08}} & \textit{0.17} & \textit{0.50} & \textit{0.50} & \textit{0.50} & \textit{0.50} \\
\midrule
\multirow{6}{*}{\rotatebox{90}{\textbf{Diagnosis}}}
& Gynecology & \best{0.00} & \best{0.00} & -0.50 & -0.17 & -0.17 & \best{0.00} & \best{0.00} & \best{0.00} & \best{0.00} & \best{0.00} \\
& Internal Med. & \best{0.00} & \best{0.00} & -0.50 & -0.33 & -0.33 & \best{0.00} & \best{0.00} & \best{0.00} & \best{0.00} & \best{0.00} \\
& Neurology & -0.17 & \best{0.00} & -0.83 & -0.33 & -0.33 & -0.17 & \best{0.00} & \best{0.00} & \best{0.00} & \best{0.00} \\
& Pediatrics & \best{0.00} & \best{0.00} & -0.67 & -0.33 & -0.33 & \best{0.00} & \best{0.00} & \best{0.00} & \best{0.00} & \best{0.00} \\
& Surgery & \best{0.00} & \best{0.00} & -0.50 & -0.17 & \best{0.00} & \best{0.00} & 0.33 & 0.25 & 0.25 & 0.25 \\
\cmidrule{2-12}
& \textit{Overall} & \textit{\best{0.00}} & \textit{\best{0.00}} & \textit{-0.50} & \textit{-0.33} & \textit{-0.33} & \textit{\best{0.00}} & \textit{\best{0.00}} & \textit{\best{0.00}} & \textit{\best{0.00}} & \textit{\best{0.00}} \\
\bottomrule
\end{tabular}
\end{table}

Fig.~\ref{fig:2}b and Table~\ref{tab:specialty_results_with_total} show the evaluation results of models acting as clinical judges, measured by the median relative error (interquartile range) between the model's consensus score and human expert consensus. For diagnostic assessment, GPT-5.1 and Gemini 3.1 achieved a median error of 0.00 (IQR: -0.50--0.10 and 0.00--0.50, respectively), while GPT-5-mini exhibited more pronounced systematic underestimation (median error: -0.50, IQR: -1.10--0.00). The on-device gpt-oss-120b demonstrated comparable alignment, achieving a median diagnostic error of 0.00 (IQR: -0.50--0.25). Notably, all three Qwen3.5 models and Gemma 4 31B also achieved a median diagnostic error of 0.00, matching the best-performing proprietary models in alignment with human expert scores. For treatment-related judgments, the gpt-oss models showed the lowest median errors (gpt-oss-20b: 0.08, IQR: -0.17--0.50; gpt-oss-120b: 0.17, IQR: -0.17--0.50), while GPT-5-mini (0.25), DeepSeek-R1 (0.29), and GPT-5.1 (0.33) showed moderate overestimation. The Qwen3.5 and Gemini models showed higher treatment errors (median 0.50), reflecting the open-ended and subjective nature of treatment evaluation. Gemma 4 31B showed a median treatment error of 0.50 (IQR: 0.00--0.75), comparable to the Qwen3.5 and Gemini models.

Model reliability was further assessed using standard agreement and stability metrics.
Inter-model agreement on clinical judgment tasks, measured using linear weighted Cohen’s $\kappa$, was moderate for diagnostic scoring ($\kappa_w \approx 0.29$--$0.74$) and lower for treatment evaluation ($\kappa_w \approx 0.11$--$0.45$), reflecting the open-ended and subjective nature of clinical management decisions.
Intra-model stability across repeated inference runs was high. For diagnostic judgment tasks, the Intraclass Correlation Coefficient (ICC) indicated excellent reliability across all models (ICC $\approx 0.82$--$0.99$), while treatment scoring showed greater variability (ICC $\approx 0.70$--$0.99$), with proprietary models exhibiting higher stability than open and on-device models.
Together, these results support the use of consensus-based aggregation to mitigate stochastic variability and produce reliable clinical outputs.

\subsection*{Fine-tuning substantially improves the diagnostic capability of on-device LLMs}
To assess the adaptability of on-device LLMs for clinical decision support, we fine-tuned the gpt-oss-20b model on the general disease diagnosis dataset (Methods). As shown in Fig.~\ref{fig:2}c (Supplementary Fig. 1--3), the fine-tuned model produced a marked improvement in diagnostic accuracy, increasing from 80.7\% (95\% CI: 74.8--85.5\%) in the base model to 86.5\% (95\% CI: 81.1--90.5\%) after adaptation. This performance exceeded the accuracy of DeepSeek-R1 (81.6\%, 95\% CI: 75.8--86.3\%) and GPT-5-mini (84.5\%, 95\% CI: 79.0--88.8\%) and approached that of GPT-5.1 (89.4\%, 95\% CI: 84.4--92.9\%).

In parallel, fine-tuning the Qwen3.5-35B model using reasoning data curated by the larger Qwen3.5-122B yielded a substantial improvement from 84.1\% (95\% CI: 78.5--88.4\%) to 87.9\% (95\% CI: 82.8--91.7\%), a gain of 3.8 percentage points (Fig.~\ref{fig:2}c). This fine-tuned accuracy closely rivals GPT-5.1 (89.4\%; $p>0.05$) and surpasses both GPT-5-mini and DeepSeek-R1, demonstrating that the family-matched training data curation strategy generalizes across model architectures.

To illustrate the impact of fine-tuning on both diagnostic accuracy and reasoning quality, we 
present three case studies evaluated on both fine-tuned models (Supplementary Cases 1--6). 
For gpt-oss-20b (Cases 1--3), Case 1 (Von Hippel-Lindau syndrome) shows that fine-tuning 
enhances reasoning quality even when the base model reaches the correct diagnosis: the base 
model provided superficial comparisons with uncertain statements, while the fine-tuned model 
employed systematic multi-system integration with detailed syndrome comparison and definitive 
synthesis. Cases 2--3 (Primary Cardiac Lymphoma and NICE lesions) demonstrate outright error 
correction, where the base model selected incorrect diagnoses and the fine-tuned model 
identified critical discriminating features that were missed. For Qwen3.5-35B (Cases 4--6), 
the base model reached correct answers across all three cases but via verbose, unfocused 
reasoning. Fine-tuning consistently transformed this into concise, structured diagnostic 
reasoning, directly identifying key discriminating features and applying systematic elimination 
rather than relying on memorised pattern-matching.

Furthermore, the sub-speciality-wise analysis (Fig.~\ref{fig:2}d) reveals that fine-tuning effectively mitigates domain-specific weaknesses observed in base models, particularly in specialized areas such as cardiovascular and breast imaging.

\begin{figure}[ht]
\centering
\includegraphics[width=0.85\textwidth]{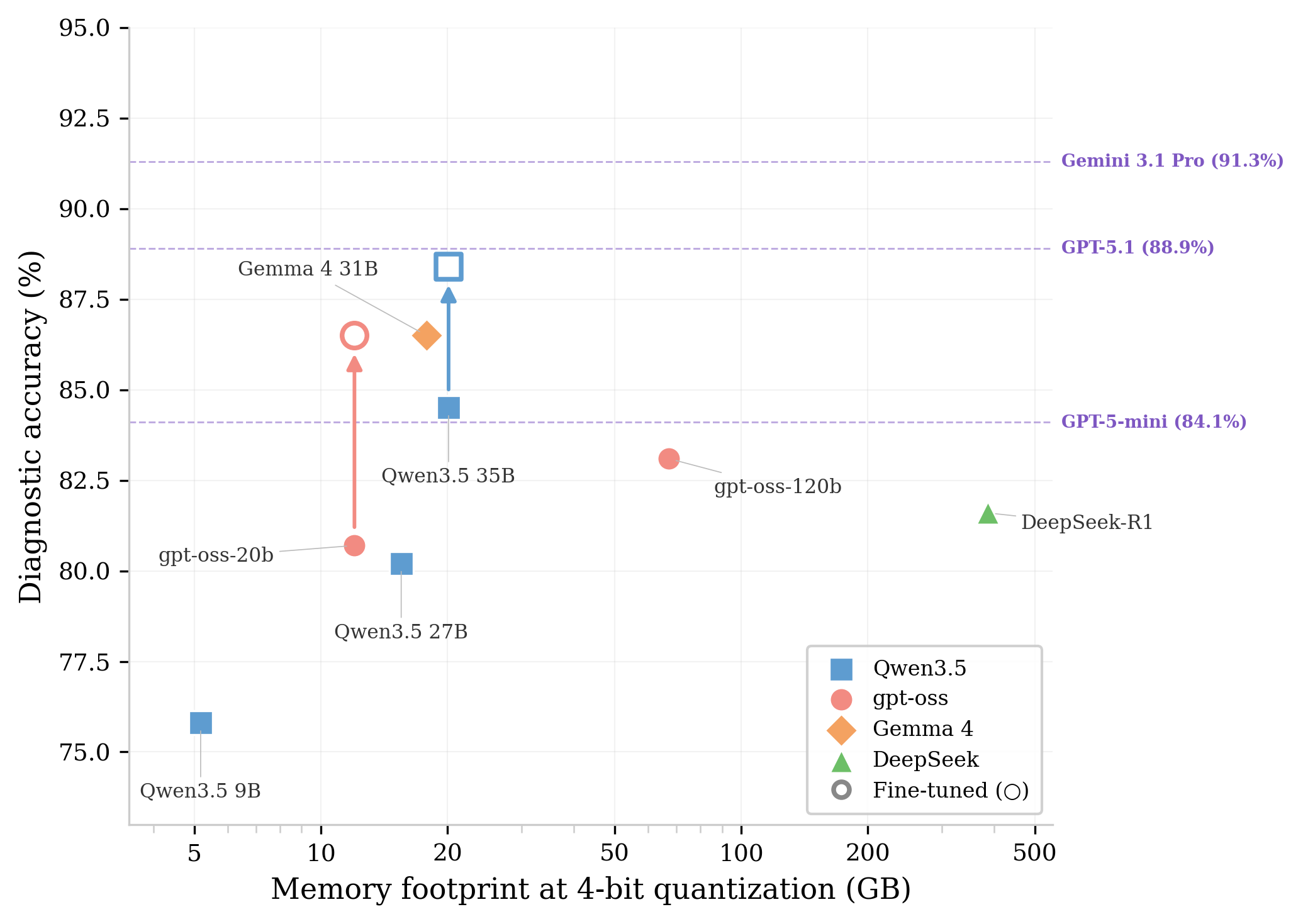}
\caption{\textbf{Parameter efficiency of open-weight models on the Eurorad general diagnosis task (N=207).}
Solid markers indicate base models; hollow markers connected by upward arrows indicate fine-tuned variants of the adjacent base model (gpt-oss-20b and Qwen3.5-35B-A3B). Diagnostic accuracy is plotted against memory footprint at 4-bit quantization (computed as parameter count $\times$ 0.5 bytes $\times$ 1.15 framework overhead factor; KV cache beyond framework defaults is not included). Parameter counts (total/active for mixture-of-experts architectures) are listed in Methods. Proprietary models (Gemini 3.1 Pro, GPT-5.1, GPT-5-mini) are shown as horizontal dashed reference lines at their observed accuracy; parameter counts and serving configurations have not been publicly disclosed and are omitted from the X axis. All accuracy values are $k=3$ majority consensus, identical to Table~\ref{tab:generalist_results}. DeepSeek-R1 is shown for reference; its memory footprint exceeds the single-GPU deployment scenario targeted by the other on-device models.}
\label{fig:3}
\end{figure}

The relationship between model size and diagnostic accuracy is summarized in Fig.~\ref{fig:3}, which plots each open-weight model's accuracy against its estimated memory footprint at 4-bit quantization. Among base models, Gemma~4 31B achieved the highest on-device accuracy (86.5\%) at approximately 18~GB, exceeding GPT-5-mini (84.5\%) while requiring only a single consumer GPU. The Qwen3.5 family showed a clear scaling trend from 9B (76.3\%, ${\sim}$5~GB) through 27B (80.2\%, ${\sim}$16~GB) to 35B (84.1\%, ${\sim}$20~GB). Fine-tuning shifted both gpt-oss-20b and Qwen3.5-35B upward without increasing their memory footprint, bringing the fine-tuned Qwen3.5-35B (87.9\%) within 1.5 percentage points of GPT-5.1 (89.4\%). By contrast, DeepSeek-R1, the strongest open-source baseline (81.6\%), requires an estimated 386~GB at 4-bit precision---well beyond single-GPU deployment. These results demonstrate that on-device models in the 5--20~GB range can match or exceed proprietary-model accuracy when combined with targeted fine-tuning, at memory footprints compatible with consumer hardware.

The substantial performance gains across both model families demonstrate that targeted domain adaptation can effectively compensate for the smaller scale of on-device models. The fine-tuned gpt-oss-20b and Qwen3.5-35B models exhibit diagnostic accuracy comparable to significantly larger proprietary systems while retaining the advantages of local deployability and privacy preservation. These findings highlight the strong potential of lightweight, fine-tuned on-device LLMs to provide high-quality clinical decision support in settings with limited computational resources.

\subsection*{Error characterization across model families}

To understand the nature of residual diagnostic errors, we classified all 358 incorrect majority-vote predictions across the 11 evaluated models into five categories using an independent LLM judge (Table~\ref{tab:error_taxonomy}; Methods). The dominant error mode was category~(c), plausible differential selection, accounting for 87.2\% of all errors (312/358). This proportion was consistent across model families: 97.2\% for proprietary models, 89.5\% for DeepSeek-R1, 83.6\% for base on-device models, and 87.0\% for fine-tuned on-device models. Category~(a) surface-form mismatches---cases where the model identified the correct diagnosis but used an alternative name or abbreviation---comprised only 5.0\% of errors (18/358), concentrated in four ground-truth labels with long multi-qualifier names. This indicates that exact-match scoring does not substantially underestimate diagnostic accuracy. Off-topic or hallucinated predictions (category~d) were restricted to the two smallest on-device models (Qwen3.5 9B and 27B), accounting for all 14 such errors; every other model produced zero off-topic errors, including all on-device models with $\geq$31B parameters as well as gpt-oss-20b (which, despite its 20B total size, activates only 3.6B parameters per token via its mixture-of-experts architecture).

We further quantified the gap between observed majority-vote accuracy and an upper-bound accuracy defined as the fraction of cases for which at least one of the three independent runs produced the correct diagnosis. This comparison separates two distinct failure modes: cases the model never answers correctly (a ceiling on generation) versus cases answered correctly in some runs but missed by majority vote (a ceiling on answer selection); the gap indicates how much accuracy could be recovered by better aggregation or verifier-based reranking without changing the underlying model. This represents the theoretical ceiling achievable by any answer-selection strategy applied to the same three generations, and is equivalent to pass@3 in the machine-learning literature~\cite{chen2021codex}. Upper-bound accuracy exceeded majority-vote accuracy by 0--11 cases per model (Supplementary Table~4). The fine-tuned Qwen3.5-35B exhibited the highest upper-bound accuracy (93.2\%, 193/207), indicating substantial recoverable performance through improved sampling or aggregation strategies. By contrast, Gemma~4 31B produced identical predictions across all three runs for every case (upper-bound gap~=~0), reflecting fully deterministic generation under our inference protocol.

To identify the boundary of current model capabilities, we defined \textit{hard-for-all} cases as those answered correctly by $\leq$36\% of models (bottom decile). This yielded 23 cases spanning all major subspecialties in rough proportion to section size, indicating that the shared knowledge frontier is distributed across the benchmark rather than concentrated in a single domain. These cases were characterized by rare histopathological subtypes (e.g., retroperitoneal ganglioneuroblastoma-intermixed, chondroblastic osteosarcoma, rosette-forming glioneuronal tumour), uncommon infections (e.g., \textit{Mycobacterium kansasii} tenosynovitis), and rare syndromes (e.g., chorea-acanthocytosis, MASA syndrome). One case (clear cell soft tissue sarcoma with melanocytic differentiation) contained three category~(a) surface-form predictions that would be scored correct under a relaxed matcher, which would remove it from the hard-for-all set; the remaining 22 cases are unambiguously hard. No single model family dominated the recoveries among these cases.

Finally, we examined the mechanism by which fine-tuning improves accuracy. We identified 19 \textit{capability-gap} cases---those answered correctly by proprietary or fine-tuned models but missed by the majority of base on-device models. Of the 13 cases that the fine-tuned Qwen3.5-35B recovered relative to its base variant, 11 (84.6\%) fell within this capability-gap set. The two cases uniquely solved by the fine-tuned model (both rare musculoskeletal entities) had analogous cases in the fine-tuning corpus: the training set contained a prior chondroblastic osteosarcoma case and related mycobacterial tenosynovitis cases from the historical Eurorad archive. This suggests that fine-tuning recovers difficult cases primarily through exposure to similar rare entities in the curated training data rather than through emergent reasoning capabilities.

\begin{table}[!htbp]
\centering
\scriptsize
\setlength{\tabcolsep}{3pt}
\caption{\textbf{Error taxonomy of incorrect majority-vote predictions on the Eurorad benchmark (N=207 cases).}
Each incorrect prediction was classified into one of five categories by an independent LLM judge (Methods). Values show the percentage of each model's total errors. Categories: (a)~surface-form mismatch, (b)~same disease family, (c)~plausible differential, (d)~off-topic/hallucinated, (e)~empty/refusal.}
\label{tab:error_taxonomy}
\begin{tabular}{l ccc @{\hskip 6pt} c @{\hskip 6pt} cc @{\hskip 6pt} cccc @{\hskip 6pt} c}
\toprule
& \multicolumn{3}{c}{\textbf{Proprietary}} & \textbf{Open} & \multicolumn{2}{c}{\textbf{gpt-oss}} & \multicolumn{4}{c}{\textbf{Qwen3.5}} & \textbf{Gemma 4} \\
\cmidrule(lr){2-4} \cmidrule(lr){5-5} \cmidrule(lr){6-7} \cmidrule(lr){8-11} \cmidrule(lr){12-12}
\textbf{Category (\%)} &
\textbf{GPT-5.1} & \textbf{GPT-5-mini} & \textbf{Gemini-3.1} & \textbf{DS-R1} &
\textbf{20b} & \textbf{120b} &
\textbf{9B} & \textbf{27B} & \textbf{35B} & \textbf{35B-FT} & \textbf{31B} \\
\midrule
(a) Surface mismatch  & 0.0  & 3.1  & 0.0   & 5.3  & 7.5  & 5.7  & 6.1  & 7.3  & 3.0  & 8.7  & 3.6 \\
(b) Same family       & 0.0  & 3.1  & 0.0   & 2.6  & 7.5  & 2.9  & 4.1  & 2.4  & 6.1  & 4.3  & 3.6 \\
(c) Plausible diff.   & 100.0 & 93.8 & 100.0 & 89.5 & 85.0 & 91.4 & 75.5 & 73.2 & 90.9 & 87.0 & 92.9 \\
(d) Off-topic$^\dagger$ & 0.0  & 0.0  & 0.0   & 0.0  & 0.0  & 0.0  & \textbf{14.3} & \textbf{17.1} & 0.0  & 0.0  & 0.0 \\
(e) Empty/refusal     & 0.0  & 0.0  & 0.0   & 2.6  & 0.0  & 0.0  & 0.0  & 0.0  & 0.0  & 0.0  & 0.0 \\
\midrule
\textbf{Total errors} & \textbf{22} & \textbf{32} & \textbf{17} & \textbf{38} & \textbf{40} & \textbf{35} & \textbf{49} & \textbf{41} & \textbf{33} & \textbf{23} & \textbf{28} \\
\bottomrule
\multicolumn{12}{l}{\footnotesize $^\dagger$Category (d) errors were produced exclusively by Qwen3.5 9B and 27B; all other models produced zero.}
\end{tabular}
\end{table}

\section*{Methods}

\subsection*{Dataset curation and pre-processing}
Case reports for the LLM-as-a-generalist task were curated from the European Society of Radiology (Eurorad)~\cite{kottlors2025eurorad}  (\url{https://www.eurorad.org/}) database across a broad spectrum of subspecialties, including musculoskeletal, cardiovascular, abdominal, uroradiology, neuroradiology, paediatric, head and neck, breast, and chest imaging.
To strictly mitigate data leakage, the test set was restricted to cases published in 2025, postdating the training cut-off of the evaluated models. From an initial pool of 350 prospective test cases, we employed GPT-5.1 to screen and exclude reports containing explicit diagnostic disclosures in the description. Historical cases published prior to 2025 were allocated for fine-tuning, yielding a final dataset comprising 1,895 training cases and 207 independent test cases.

Cases for the LLM-as-a-specialist task were based on the recent ophthalmology multiple-choice question dataset~\cite{ophthalmologyQA}. It contains 130 questions covering six topics: anterior segment diseases, external eye/orbital diseases, glaucoma, ocular trauma, refractive disorders/strabismus, and retinal diseases. Each question has one to six correct answers from five to nine answer choices.

Cases for the LLM-as-a-clinical-judge scoring task were curated from the benchmark data in~\cite{eval-deepseek125Patients}, comprising 125 patient cases. 
Each case included a chief complaint and up to five diagnostic or treatment recommendations generated by distinct models, including GPT-4, GPT-4o, GPT-3.5, Gem2FTE, and DeepSeek-R1. In this task, the evaluated LLMs were required to audit these predictions by assigning quality scores rather than generating de novo diagnoses. Performance was assessed by measuring the concordance between model-generated scores and reference ratings provided by medical experts.

\subsection*{Task-specific inference protocols}

To ensure fair comparison, we developed standardized zero-shot inference pipelines for each clinical task. All protocols utilized a self-consistency framework where each case was queried three independent times (\(k = 3\)).

\textbf{General Radiology Diagnosis.}
The LLM-as-a-generalist diagnostic task was formulated as a constrained single-label selection problem (Supplementary Prompt 1). Models were provided with the patient history and imaging findings and instructed to select the most likely diagnosis strictly verbatim from a provided differential list. Post-processing utilized a deterministic regex-based extractor to isolate the final diagnosis from the generated reasoning stream.

\textbf{Ophthalmology Specialty QA.}
The LLM-as-a-specialist task involved complex multiple-choice questions (MCQs) requiring multi-label classification. Unlike the radiology task, models were instructed to ``Select ALL correct answers'' from options A–Z. The system prompt enforced a strict output format consisting only of concatenated capital letters (e.g., ``ABE'' or ``D''), prohibiting explanatory text in the final output to facilitate automated parsing  (Supplementary Prompt 2).

\textbf{Clinical Judgment Simulation.}
The LLM-as-a-clinical-judge task evaluated the models' ability to simulate expert clinical judgment. Models were presented with a clinical case adapted from ~\cite{eval-deepseek125Patients}, alongside a candidate diagnosis/treatment plan, and a reference standard. They were instructed to assign a quality score on a 5-point Likert scale (1 = Most relevant options missing, 5 = All relevant options mentioned) based on a strict scoring rubric. Half-point scores (e.g., 4.5) were permitted to capture granular distinctions in quality  (Supplementary Prompt 3).

\subsection*{Inference Protocol for Zero-shot Experiments}

We benchmarked three distinct categories of models to represent the current landscape of LLMs:

\textbf{Proprietary Frontier Models (GPT-5.1, GPT-5-mini, and Gemini 3.1 Pro).}
We selected OpenAI’s \texttt{gpt-5.1-2025-11-13} and \texttt{gpt-5-mini} reasoning models, alongside Google’s \texttt{gemini-3.1-pro}. These models represent the current state-of-the-art in closed-source reasoning. Inference for OpenAI models was conducted via the OpenAI Responses API, while Gemini 3.1 Pro was accessed through the Google AI API. To evaluate the impact of inference-time compute on OpenAI models, we modulated the \texttt{reasoning\_effort} parameter (\texttt{low}, \texttt{medium}, \texttt{high}). For all tasks, we utilized the default \texttt{medium} effort setting.

\textbf{Open-Source State-of-the-Art Model (DeepSeek-R1).}
To represent the pinnacle of open-source capability, we evaluated the 671-billion parameter \texttt{DeepSeek-R1-0528}. This model utilizes large-scale reinforcement learning to optimize reasoning paths and is currently the strongest non-proprietary baseline available. Inference was performed via the OpenRouter API. We utilized a context window of 8{,}192 tokens to strictly enforce the capture of the model's native reinforcement-learning-aligned reasoning traces. 

\textbf{On-Device Models (gpt-oss, Qwen3.5, and Gemma 4).}
The primary focus of this study is on-device LLMs from three model families. From the \texttt{gpt-oss} family, we evaluated both the 20-billion parameter variant (optimized for consumer GPUs) and the 120-billion parameter variant. The gpt-oss models were evaluated using the Hugging Face Inference Router, targeting the Fireworks AI provider for the 20b model and Cerebras for the 120b model to maximize throughput. Similar to the proprietary OpenAI models, we modulated the \texttt{reasoning\_effort} parameter (\texttt{low}, \texttt{medium}, \texttt{high}) through the system prompt to trigger extended chain-of-thought generation patterns. From the Qwen3.5 family, we evaluated three model sizes: 9B, 27B, and 35B parameters, as well as a fine-tuned 35B variant. The Qwen3.5 models were accessed via the Hugging Face Inference Router with high reasoning effort. Output parsing relied on a custom regex extractor to identify valid answer sequences within the generated text.
From the Gemma family, we evaluated Gemma 4 31B, Google's open-weight on-device model. Inference was performed via the Hugging Face Inference Router using the same self-consistency protocol ($k=3$). Output parsing used the same regex-based extractor applied to other on-device models.

\subsection*{Error taxonomy analysis}

To characterize residual diagnostic errors on the Eurorad benchmark, we classified every incorrect $k=3$ majority-vote prediction into five categories: (a)~surface-form mismatch (clinically correct but differing in wording, synonym, or abbreviation), (b)~same disease family (correct organ system or disease group, wrong specific entity), (c)~plausible differential (a diagnosis on the provided list, but incorrect), (d)~off-topic (not on the differential list, anatomically unrelated, or hallucinated), and (e)~empty or refusal (no diagnosis produced or parse failure). Classification was performed by Claude Opus 4.6 (Anthropic) via AWS Bedrock, selected as a neutral judge external to the evaluation set to eliminate self-evaluation bias. For each error, the judge received the clinical history, differential diagnosis list, ground-truth diagnosis, and model prediction, and was instructed to assign exactly one category with a one-sentence justification. The full judge prompt is provided in Supplementary Prompt 6. Case-level difficulty was computed as the fraction of models answering correctly per case, cross-tabulated by model family (proprietary, open-source large, on-device base, on-device fine-tuned). Hard-for-all cases were defined as the bottom decile ($\leq$36\% of models correct). Capability-gap cases were defined as those answered correctly by proprietary or fine-tuned models but missed by the majority of base on-device models. Upper-bound accuracy (equivalent to pass@3 in the machine-learning literature~\cite{chen2021codex}) was computed per model as the fraction of cases where at least one of the three runs matched the ground truth.

\subsection*{Training Dataset Preparation for Radiological Cases}

The gpt-oss family consists of reasoning-capable language models designed to generate structured Chain-of-Thought (CoT) reasoning during inference. To develop high-quality training data for fine-tuning gpt-oss-20b on radiological diagnosis tasks, we employed gpt-oss-120b---a substantially larger model from the same architectural family---to curate systematic diagnostic reasoning data for all 1,895 cases in the Eurorad dataset. The use of gpt-oss-120b for data curation was predicated on three key advantages: (i) architectural consistency between the 120b and 20b variants ensures compatibility of reasoning patterns, (ii) the larger parameter count enables more sophisticated medical reasoning and systematic differential diagnosis evaluation, and (iii) automated generation provides scalable, consistent reasoning data across the entire dataset.

For each radiological case, gpt-oss-120b was provided with the clinical case presentation (comprising patient history and imaging findings), the original expert radiologist discussion from the Eurorad dataset as contextual grounding, and the list of differential diagnoses. The original discussion, which provided academic descriptions of case characteristics and imaging findings, served as reference material to inform the generation of improved systematic reasoning. The model was instructed to generate Chain-of-Thought reasoning following a structured four-step diagnostic framework: (1) symptom-finding correlation---establishing connections between clinical presentation and imaging observations, (2) differential mapping---evaluating how imaging findings support or contradict each candidate diagnosis, (3) systematic elimination---providing explicit reasoning for excluding less likely diagnostic options, and (4) diagnostic convergence---demonstrating the logical pathway to the final diagnosis. Reasoning generation employed the following parameters: temperature = 0.6, maximum output tokens = 2000, target length = 200--400 words, accessed via the HuggingFace Inference API.

Generated reasoning samples underwent systematic validation to ensure data quality. We verified that all 1,895 generated reasoning chains converged to the correct ground truth diagnosis, confirming alignment between the model's reasoning process and the clinically validated diagnoses. Automated quality metrics assessed each response for: (i) appropriate length (200--600 words acceptable range), (ii) presence of all four required reasoning components (symptom-finding correlation, differential mapping, systematic elimination, and diagnostic convergence), and (iii) response completeness (absence of premature truncation). Additionally, a subset of cases was manually inspected to validate that the systematic elimination reasoning across all four reasoning phases logically progressed toward the correct diagnosis, and that the final diagnosis selection matched the ground truth. For this subset, the generated reasoning was also compared against the original discussions to ensure clinical accuracy and logical coherence. The complete dataset comprising all 1,895 cases was used for gpt-oss-20b fine-tuning.

\subsection*{Training Protocol}
Fine-tuning of the gpt-oss-20b model was performed using a curated reasoning dataset generated by its larger counterpart, gpt-oss-120b. This approach ensured that the smaller model learned from systematic diagnostic patterns that were architecturally compatible with its native reasoning framework. The model was loaded using the Unsloth framework with 4-bit quantization to enable efficient training on limited computational resources. A maximum sequence length of 4,096 tokens was configured to accommodate the clinical case presentations and associated reasoning chains.

Parameter-efficient fine-tuning was implemented using Low-Rank Adaptation (LoRA) with rank r=32 and alpha=64, targeting all linear layers in the model architecture, including query, key, value, and output projections, as well as the feed-forward network components (gate, up, and down projections) across all 32 transformer layers~\cite{hu2022lora, gao2025loraperiop,le2025impact}. Additionally, expert layers within the mixture-of-experts architecture were targeted at strategic depths: early layers (0-7) for initial processing, middle layers (8-15) for pattern recognition and reasoning, upper layers (16-23) for deep reasoning, and deep layers (24-31) for final refinement and output generation. This comprehensive targeting strategy ensured that the model could effectively learn diagnostic reasoning patterns at multiple levels of abstraction. A LoRA dropout rate of 0.05 was applied to prevent overfitting, and gradient checkpointing was enabled to reduce memory consumption during training.

The curated reasoning data was formatted using the gpt-oss chat template with medium reasoning effort, incorporating the gpt-oss-120b-generated reasoning as structured thinking content. This approach enabled the model to learn from the systematic diagnostic patterns demonstrated by the larger model while maintaining compatibility with the gpt-oss reasoning framework. Training was conducted over 3 epochs using the AdamW optimizer with a learning rate of $1 \times 10^{-4}$, cosine learning rate scheduling, and a warmup ratio of 0.1. 

In parallel, fine-tuning of the Qwen3.5-35B-A3B model was conducted to evaluate the adaptability of the Qwen family. Because different model families exhibit distinct intrinsic reasoning styles, we curated a separate reasoning dataset generated specifically by the larger Qwen3.5-122B model to maintain alignment and prevent formatting clashes during training. To ensure optimal training stability, the Qwen3.5 model was loaded in 16-bit precision (bfloat16) without 4-bit quantization, in accordance with Unsloth documentation advising against MoE QLoRA for this specific architecture. The maximum sequence length was adjusted to 2,048 tokens to safely manage memory scaling constraints. LoRA was applied globally to key attention and feed-forward modules with a rank of r=16 and alpha=32. A LoRA dropout rate of 0 was strictly enforced to maintain compatibility with the MoE expert layer parameter wrappers, alongside Unsloth-optimized gradient checkpointing. The data was formatted using the Qwen3.5 chat template, with structured diagnostic reasoning steps embedded explicitly within <think> tags. Training utilized the 8-bit AdamW optimizer over 3 epochs with a reduced learning rate of 5e-5, maintaining the cosine scheduling and 0.1 warmup ratio.

For both models, mixed-precision training with bfloat16 was utilized to accelerate computation while maintaining numerical stability, and all experiments were conducted with fixed random seeds to ensure reproducibility.

\subsection*{Inference protocol of fine-tuned models}

To evaluate the fine-tuned models on radiological diagnosis, we employed tailored inference strategies to account for the differing baseline capacities of the underlying architectures. For the smaller gpt-oss-20b model, we designed a controlled inference pipeline optimized for the exploration of deterministic and diverse hypotheses. Our goal was to demonstrate that advanced inference strategies can be leveraged to further boost the performance and reliability of highly constrained models. All predictions for this model were generated using group beam search, a decoding strategy that encourages exploration across diverse reasoning paths while maintaining stability in high-stakes clinical settings. After systematic experimentation with different decoding settings, we found that a configuration of 13 beams, 13 beam groups, and a diversity penalty of 0.5 provided the strongest performance on the Eurorad validation set. This setup enforced full beam-group separation, ensuring that each beam starts its own reasoning trajectory and looks at the presented case from a different angle, which is particularly effective in reducing mode collapse and repetitive reasoning. Maximum generation length was set to 3,000 tokens to accommodate longer Chain-of-Thought outputs, and sampling was disabled to ensure reproducibility across runs.

For each case, the gpt-oss-20b model produced 13 independent reasoning traces. Final predictions were obtained using a majority-vote aggregation over the extracted diagnostic answers, with ties resolved by selecting the earliest beam. All inputs were encoded with the gpt-oss chat template using left padding and a 4,096-token context window, and inference was executed using the Unsloth runtime with 4-bit quantized weights and attached LoRA adapters. This protocol allowed the model to balance the range of diagnostic reasoning with reliability, resulting in a stable exact-match performance while preserving a clinically interpretable diagnostic rationale.

Conversely, the fine-tuned Qwen3.5-35B model demonstrated robust diagnostic reasoning without the need for additional decoding strategies. Inference was conducted using the Unsloth runtime with 4-bit quantization. Inputs were processed within a 2,048-token maximum sequence length. We employed standard autoregressive generation with a temperature of 0.7 and a maximum generation length of 4,000 tokens to safely accommodate its extensive, multi-phase chain-of-thought outputs. This contrast in inference protocols highlights that while advanced search strategies are valuable for extracting maximal performance from smaller models, other models can achieve state-of-the-art clinical accuracy using default inference configurations.

\subsection*{Statistical analysis}
To account for the stochastic nature of Large Language Model (LLM) generation, we employed a self-consistency framework wherein each model was queried three independent times ($k=3$) for every case. All performance metrics and statistical comparisons were derived from the consensus prediction of these three runs. For nominal tasks, including the general diagnosis questions and ophthalmology specialty-specific questions, the consensus was determined via majority voting, where a prediction was considered correct only if the correct answer was generated in at least two of the three runs. For ordinal tasks evaluated on a 5-point Likert scale (LLM-as-a-judge for diagnosis and treatment scoring), the consensus was defined as the mean score of the three runs to obtain a stable per-case consensus score.

Model performance for nominal tasks was reported as accuracy, with 95\% Confidence Intervals (CIs) calculated using the Wilson Score Interval method to provide robust estimates for binomial proportions. For the clinical judge task, performance was reported as the signed median error across cases between the model’s mean consensus score and the ground-truth human expert score, summarized using the interquartile range (IQR). Statistical significance between model performances was determined using pairwise hypothesis tests on paired samples evaluated on the same test cases, with a significance threshold of $P < 0.05$. Differences in accuracy for nominal tasks were assessed using McNemar’s Test with continuity correction, while differences in the distribution of errors for ordinal clinical judgment tasks were assessed using the Wilcoxon Signed-Rank Test.

We further evaluated both the internal stability of the models and agreement between models using metrics appropriate to the nominal and ordinal structure of the evaluation tasks. To measure generation consistency across the three inference runs (intra-model stability), we calculated Fleiss’ Kappa ($\kappa$) for nominal datasets and the Intraclass Correlation Coefficient (ICC, form 3,k) for ordinal datasets. To assess the degree to which different models converged on identical predictions independent of ground truth (inter-model agreement), we calculated standard Cohen’s Kappa for nominal tasks. For ordinal LLM-as-a-judge scoring tasks, Linear Weighted Kappa was employed to penalize partial disagreements (e.g., scores of 4 vs. 5) less severely than complete disagreements.

\section*{Discussion}

This study systematically evaluates the capabilities of on-device large language models across three representative clinical tasks: general diagnosis, specialty-level reasoning, and simulation of expert judgment. Across all settings, the gpt-oss, Qwen3.5, and Gemma 4 models demonstrate clinically meaningful performance despite their substantially smaller scale relative to frontier proprietary models (such as GPT-5.1, GPT-5-mini, and Gemini 3.1 Pro), underscoring the feasibility of deploying lightweight LLMs directly within healthcare institutions.

In the general disease diagnosis task, on-device models achieved strong zero-shot performance, showing that compact architectures can capture the broad clinical reasoning patterns required for diagnosis. Notably, the Qwen3.5-35B base model matched the proprietary GPT-5-mini (84.1\% vs 84.5\%), while the 27B variant (80.2\%) performed comparably to DeepSeek-R1 (81.6\%), demonstrating that multiple on-device architectures can independently achieve strong clinical reasoning. Gemma 4 31B further extended this pattern, achieving the strongest base on-device accuracy (86.5\%) without fine-tuning and exceeding GPT-5-mini. The consistency of strong performance across three independent model families suggests that the viability of locally deployable clinical decision support is not contingent on a specific architecture. Moreover, in the ophthalmology specialty task, the on-device gpt-oss-120b model outperformed both GPT-5.1 and GPT-5-mini while ranking third overall behind Gemini 3.1 Pro and DeepSeek-R1. This is noteworthy given that most existing clinical LLM evaluations have focused on large cloud-based models~\cite{zhang2025ophthreview, bhayana2024radiology, schramm2025multimodal}, leaving open the question of whether smaller locally deployable systems can achieve comparable levels of generalization. Our findings indicate that modern on-device architectures can support robust performance in common diagnostic scenarios without requiring external computation or data transfer.

The ability of LLMs to act as "clinical judges" is critical for scalable quality assurance and automated evaluation~\cite{kocaman2025clever, croxford2025evaluating}. Our results indicate that on-device models can align closely with human expert consensus, often exhibiting greater stability than efficient proprietary alternatives like GPT-5-mini. All three Qwen3.5 models achieved median diagnostic errors of 0.00, matching the best-performing models and confirming that lightweight architectures can reliably simulate expert judgment. This suggests that compact models can support evaluative tasks that require not only factual knowledge but also sensitivity to clinical reasoning standards and rubric-based assessment criteria. This capability is essential for deploying LLMs as automated auditors or second-opinion systems in clinical workflows~\cite{humaneval2025llmjudge, genovese2026artificial}.

Beyond zero-shot performance, the adaptability of on-device models represents a substantial advantage over closed-source systems. Fine-tuning the gpt-oss-20b and Qwen3.5-35B models on general diagnosis data substantially improved their performance, with the fine-tuned Qwen3.5-35B achieving an exact match accuracy of 87.9\% (95\% CI: 82.8--91.7\%), closely rivaling GPT-5.1 (89.4\%). We observed that fine-tuning yielded the most robust improvements when models were trained on reasoning traces generated by a larger, architecturally matched model from the same family (e.g., utilizing Qwen-122B for the Qwen3.5 series, and gpt-oss-120b for the gpt-oss series). Because different model families exhibit intrinsic variations in how they structure logic, this family-matched approach preserves the native reasoning style, preventing formatting clashes and optimizing the assimilation of systematic diagnostic patterns. This result suggests that domain-specific optimization, when aligned with a model's inherent reasoning architecture, can efficiently compensate for smaller model scale~\cite{davis2026medslice, guluzade2025elmtex, wind2025rar}. As illustrated in Fig.~\ref{fig:3}, both fine-tuned models close most of the gap to GPT-5.1 without any increase in memory footprint, demonstrating that accuracy gains from fine-tuning are orthogonal to model size. Fine-tuning also enhanced robustness across radiology subspecialties, yielding more uniform performance and mitigating weaknesses observed in the base models. These findings highlight that clinics can deploy efficient, low-cost, customized AI tools tailored to their specific patient demographics and disease prevalences without compromising data privacy.

Taken together, the experiments in this study highlight three key insights. First, compact on-device LLMs across multiple model families (gpt-oss, Qwen3.5, Gemma 4) can provide strong general diagnostic reasoning and domain-specific performance. Second, on-device LLMs can approximate expert judgment with surprising fidelity, positioning them as valuable components of locally governed AI ecosystems that support both clinical decision-making and meta-evaluative tasks. Third, fine-tuning plays a critical role in achieving competitive accuracy across diverse subspecialties, allowing healthcare institutions to develop tailored high-performing models from relatively small architectures.

Error characterization of incorrect predictions reveals a specific failure profile that further contextualizes these findings. Across all model families, the overwhelming majority of errors (87.2\%) are plausible differential selections---the model selects another diagnosis from the provided list rather than hallucinating or producing an off-topic response. This indicates that even when models err, they are performing constrained clinical reasoning within the differential. Off-topic or hallucinated predictions were confined to Qwen3.5 9B and 27B; all other evaluated models, including both fine-tuned variants, produced zero such errors. Even for the affected models, the absolute rate was low ($\sim$3.4\% of cases), suggesting that hallucination risk is bounded rather than pervasive at this scale. Fine-tuning preferentially recovers capability-gap cases---those solvable by larger models but missed by base on-device variants---with 84.6\% of the fine-tuned Qwen3.5-35B's recovered cases falling within this set. The 23 hard-for-all cases that remain unsolved by most models are distributed proportionally across subspecialties and consist of rare histopathological subtypes and uncommon syndromes, defining a shared knowledge frontier that reflects the long tail of medical knowledge rather than a systematic architectural limitation.

Our study also has limitations. First, while our benchmarks cover diagnosis, management, and evaluation, they rely on retrospective patient cases and examination questions, which may not fully capture the complexity and noise of real-time clinical environments. The error taxonomy analysis indicates that 5.0\% of incorrect predictions (18 of 358 errors) are surface-form mismatches (category~a), i.e., clinically correct diagnoses scored as errors under exact-match evaluation, suggesting that the reported accuracies conservatively estimate diagnostic capability. The performance gains from fine-tuning partly reflect exposure to similar rare entities in the historical Eurorad training corpus; disentangling pattern recall from improved reasoning capability would require held-out rare-entity test cases absent from the training data, which we identify as future work. Human validation of a random subsample of the LLM-judge error classifications was not completed for this submission; the error category proportions reported here depend on the judge's consistency, which we assess through manual review of category~(a) assignments (Results). Local deployment mitigates privacy and data governance concerns but introduces operational challenges, including hardware reliability, secure integration with clinical information systems, and ongoing monitoring of model behavior~\cite{asgari2025hallucination, omar2025adversarial}. Fine-tuning requires access to high-quality labeled data, which may be limited in certain specialties. Building on our findings regarding model-specific logic patterns, part of our future work will explore the variations in intrinsic reasoning styles between different vendors. Specifically, investigating multi-vendor, agentic debate frameworks—where diverse models collaborate, critique, or iteratively refine each other's diagnostic reasoning—could provide a novel mechanism to further enhance accuracy and reliability in highly complex clinical scenarios.

In summary, this work demonstrates that on-device LLMs offer a promising and practical alternative to large proprietary systems for clinical decision support. These compact models can achieve reliable diagnostic reasoning, robust subspecialty performance, and alignment with expert judgment---all while maintaining strict control over patient data and computational infrastructure. These features position on-device LLMs as strong candidates for safe, scalable, and equitable integration of AI into clinical practice.

\subsection*{Data Availability}
The benchmarking results and model outputs generated in this study are available in the Supplementary Information. The raw input data for the general diagnosis task are available from the Eurorad library (\url{https://www.eurorad.org/}); a script to retrieve the specific cases used in this study is provided in the code repository. The ophthalmology and clinical judge datasets are publicly available at \url{https://github.com/bowang-lab/on-device-LLM}. The training dataset with gpt-oss-120b reasoning enhancement used to fine-tune the model is available at \url{https://huggingface.co/datasets/wanglab/eurorad-gpt-oss-training-data}.

\subsection*{Code Availability}
The code for model inference, benchmarking, and evaluation is publicly available on GitHub at \url{https://github.com/bowang-lab/on-device-LLM}. The fine-tuned model weights are available on HuggingFace at \url{https://huggingface.co/wanglab/on-device-LLM-gpt-oss-20b}.

\subsection*{Acknowledgements} 
This work was supported by the Natural Sciences and Engineering Research Council of Canada (RGPIN-2020-06189 and DGECR-2020-00294) and CIFAR AI Chair programs. This research was enabled, in part, by computing resources provided by the Digital Research Alliance of Canada.


\bibliographystyle{IEEEtran}
\bibliography{main-ref}

@article{DeepSeek-R1,
  title = {{DeepSeek-R1} incentivizes reasoning in {LLMs} through reinforcement learning},
  volume = {645},
  number = {8081},
  journal = {Nature},
  author = {Guo,  Daya and Yang,  Dejian and Zhang,  Haowei and Song,  Junxiao and Wang,  Peiyi and Zhu,  Qihao and Xu,  Runxin and Zhang,  Ruoyu and Ma,  Shirong and Bi,  Xiao and Zhang,  Xiaokang and Yu,  Xingkai and Wu,  Yu and Wu,  Z. F. and Gou,  Zhibin and Shao,  Zhihong and Li,  Zhuoshu and Gao,  Ziyi and Liu,  Aixin and Xue,  Bing and Wang,  Bingxuan and Wu,  Bochao and Feng,  Bei and Lu,  Chengda and Zhao,  Chenggang and Deng,  Chengqi and Ruan,  Chong and Dai,  Damai and Chen,  Deli and Ji,  Dongjie and Li,  Erhang and Lin,  Fangyun and Dai,  Fucong and Luo,  Fuli and Hao,  Guangbo and Chen,  Guanting and Li,  Guowei and Zhang,  H. and Xu,  Hanwei and Ding,  Honghui and Gao,  Huazuo and Qu,  Hui and Li,  Hui and Guo,  Jianzhong and Li,  Jiashi and Chen,  Jingchang and Yuan,  Jingyang and Tu,  Jinhao and Qiu,  Junjie and Li,  Junlong and Cai,  J. L. and Ni,  Jiaqi and Liang,  Jian and Chen,  Jin and Dong,  Kai and Hu,  Kai and You,  Kaichao and Gao,  Kaige and Guan,  Kang and Huang,  Kexin and Yu,  Kuai and Wang,  Lean and Zhang,  Lecong and Zhao,  Liang and Wang,  Litong and Zhang,  Liyue and Xu,  Lei and Xia,  Leyi and Zhang,  Mingchuan and Zhang,  Minghua and Tang,  Minghui and Zhou,  Mingxu and Li,  Meng and Wang,  Miaojun and Li,  Mingming and Tian,  Ning and Huang,  Panpan and Zhang,  Peng and Wang,  Qiancheng and Chen,  Qinyu and Du,  Qiushi and Ge,  Ruiqi and Zhang,  Ruisong and Pan,  Ruizhe and Wang,  Runji and Chen,  R. J. and Jin,  R. L. and Chen,  Ruyi and Lu,  Shanghao and Zhou,  Shangyan and Chen,  Shanhuang and Ye,  Shengfeng and Wang,  Shiyu and Yu,  Shuiping and Zhou,  Shunfeng and Pan,  Shuting and Li,  S. S. and Zhou,  Shuang and Wu,  Shaoqing and Yun,  Tao and Pei,  Tian and Sun,  Tianyu and Wang,  T. and Zeng,  Wangding and Liu,  Wen and Liang,  Wenfeng and Gao,  Wenjun and Yu,  Wenqin and Zhang,  Wentao and Xiao,  W. L. and An,  Wei and Liu,  Xiaodong and Wang,  Xiaohan and Chen,  Xiaokang and Nie,  Xiaotao and Cheng,  Xin and Liu,  Xin and Xie,  Xin and Liu,  Xingchao and Yang,  Xinyu and Li,  Xinyuan and Su,  Xuecheng and Lin,  Xuheng and Li,  X. Q. and Jin,  Xiangyue and Shen,  Xiaojin and Chen,  Xiaosha and Sun,  Xiaowen and Wang,  Xiaoxiang and Song,  Xinnan and Zhou,  Xinyi and Wang,  Xianzu and Shan,  Xinxia and Li,  Y. K. and Wang,  Y. Q. and Wei,  Y. X. and Zhang,  Yang and Xu,  Yanhong and Li,  Yao and Zhao,  Yao and Sun,  Yaofeng and Wang,  Yaohui and Yu,  Yi and Zhang,  Yichao and Shi,  Yifan and Xiong,  Yiliang and He,  Ying and Piao,  Yishi and Wang,  Yisong and Tan,  Yixuan and Ma,  Yiyang and Liu,  Yiyuan and Guo,  Yongqiang and Ou,  Yuan and Wang,  Yuduan and Gong,  Yue and Zou,  Yuheng and He,  Yujia and Xiong,  Yunfan and Luo,  Yuxiang and You,  Yuxiang and Liu,  Yuxuan and Zhou,  Yuyang and Zhu,  Y. X. and Huang,  Yanping and Li,  Yaohui and Zheng,  Yi and Zhu,  Yuchen and Ma,  Yunxian and Tang,  Ying and Zha,  Yukun and Yan,  Yuting and Ren,  Z. Z. and Ren,  Zehui and Sha,  Zhangli and Fu,  Zhe and Xu,  Zhean and Xie,  Zhenda and Zhang,  Zhengyan and Hao,  Zhewen and Ma,  Zhicheng and Yan,  Zhigang and Wu,  Zhiyu and Gu,  Zihui and Zhu,  Zijia and Liu,  Zijun and Li,  Zilin and Xie,  Ziwei and Song,  Ziyang and Pan,  Zizheng and Huang,  Zhen and Xu,  Zhipeng and Zhang,  Zhongyu and Zhang,  Zhen},
  year = {2025},
  pages = {633–638}
}

@article{ophthalmologyQA,
title = {{DeepSeek-R1} outperforms {Gemini}~2.0~{Pro}, {OpenAI}~o1, and o3-mini in bilingual complex ophthalmology reasoning},
journal = {Advances in Ophthalmology Practice and Research},
volume = {5},
number = {3},
pages = {189-195},
year = {2025},
author = {Pusheng Xu and Yue Wu and Kai Jin and Xiaolan Chen and Mingguang He and Danli Shi}
}

@article{openai-AIConsult,
  title={{AI}-based clinical decision support for primary care: A real-world study},
  author={Korom, Robert and Kiptinness, Sarah and Adan, Najib and Said, Kassim and Ithuli, Catherine and Rotich, Oliver and Kimani, Boniface and King'ori, Irene and Kamau, Stellah and Atemba, Elizabeth and others},
  journal={arXiv preprint arXiv:2507.16947},
  year={2025}
}

@article{gpt-oss,
Author = {OpenAI and Sandhini Agarwal and Lama Ahmad and Jason Ai and Sam Altman and Andy Applebaum and Edwin Arbus and Rahul K. Arora and Yu Bai and Bowen Baker and Haiming Bao and Boaz Barak and Ally Bennett and Tyler Bertao and Nivedita Brett and Eugene Brevdo and Greg Brockman and Sebastien Bubeck and Che Chang and Kai Chen and Mark Chen and Enoch Cheung and Aidan Clark and Dan Cook and Marat Dukhan and Casey Dvorak and Kevin Fives and Vlad Fomenko and Timur Garipov and Kristian Georgiev and Mia Glaese and Tarun Gogineni and Adam Goucher and Lukas Gross and Katia Gil Guzman and John Hallman and Jackie Hehir and Johannes Heidecke and Alec Helyar and Haitang Hu and Romain Huet and Jacob Huh and Saachi Jain and Zach Johnson and Chris Koch and Irina Kofman and Dominik Kundel and Jason Kwon and Volodymyr Kyrylov and Elaine Ya Le and Guillaume Leclerc and James Park Lennon and Scott Lessans and Mario Lezcano-Casado and Yuanzhi Li and Zhuohan Li and Ji Lin and Jordan Liss and Lily (Xiaoxuan) Liu and Jiancheng Liu and Kevin Lu and Chris Lu and Zoran Martinovic and Lindsay McCallum and Josh McGrath and Scott McKinney and Aidan McLaughlin and Song Mei and Steve Mostovoy and Tong Mu and Gideon Myles and Alexander Neitz and Alex Nichol and Jakub Pachocki and Alex Paino and Dana Palmie and Ashley Pantuliano and Giambattista Parascandolo and Jongsoo Park and Leher Pathak and Carolina Paz and Ludovic Peran and Dmitry Pimenov and Michelle Pokrass and Elizabeth Proehl and Huida Qiu and Gaby Raila and Filippo Raso and Hongyu Ren and Kimmy Richardson and David Robinson and Bob Rotsted and Hadi Salman and Suvansh Sanjeev and Max Schwarzer and D. Sculley and Harshit Sikchi and Kendal Simon and Karan Singhal and Yang Song and Dane Stuckey and Zhiqing Sun and Philippe Tillet and Sam Toizer and Foivos Tsimpourlas and Nikhil Vyas and Eric Wallace and Xin Wang and Miles Wang and Olivia Watkins and Kevin Weil and Amy Wendling and Kevin Whinnery and Cedric Whitney and Hannah Wong and Lin Yang and Yu Yang and Michihiro Yasunaga and Kristen Ying and Wojciech Zaremba and Wenting Zhan and Cyril Zhang and Brian Zhang and Eddie Zhang and Shengjia Zhao},
title = {gpt-oss-120b \& gpt-oss-20b Model Card},
year = {2025},
journal = {arXiv preprint arXiv:2508.10925},
}

@article{MedPALM1,
  title = {Large language models encode clinical knowledge},
  volume = {620},
  number = {7972},
  journal = {Nature},
  author = {Singhal,  Karan and Azizi,  Shekoofeh and Tu,  Tao and Mahdavi,  S. Sara and Wei,  Jason and Chung,  Hyung Won and Scales,  Nathan and Tanwani,  Ajay and Cole-Lewis,  Heather and Pfohl,  Stephen and Payne,  Perry and Seneviratne,  Martin and Gamble,  Paul and Kelly,  Chris and Babiker,  Abubakr and Sch\"{a}rli,  Nathanael and Chowdhery,  Aakanksha and Mansfield,  Philip and Demner-Fushman,  Dina and Ag\"{u}era y Arcas,  Blaise and Webster,  Dale and Corrado,  Greg S. and Matias,  Yossi and Chou,  Katherine and Gottweis,  Juraj and Tomasev,  Nenad and Liu,  Yun and Rajkomar,  Alvin and Barral,  Joelle and Semturs,  Christopher and Karthikesalingam,  Alan and Natarajan,  Vivek},
  year = {2023},
  pages = {172–180}
}

@article{MedPaLM2,
  title = {Toward expert-level medical question answering with large language models},
  volume = {31},
  number = {3},
  journal = {Nature Medicine},
  author = {Singhal,  Karan and Tu,  Tao and Gottweis,  Juraj and Sayres,  Rory and Wulczyn,  Ellery and Amin,  Mohamed and Hou,  Le and Clark,  Kevin and Pfohl,  Stephen R. and Cole-Lewis,  Heather and Neal,  Darlene and Rashid,  Qazi Mamunur and Schaekermann,  Mike and Wang,  Amy and Dash,  Dev and Chen,  Jonathan H. and Shah,  Nigam H. and Lachgar,  Sami and Mansfield,  Philip Andrew and Prakash,  Sushant and Green,  Bradley and Dominowska,  Ewa and Ag\"{u}era y Arcas,  Blaise and Tomašev,  Nenad and Liu,  Yun and Wong,  Renee and Semturs,  Christopher and Mahdavi,  S. Sara and Barral,  Joelle K. and Webster,  Dale R. and Corrado,  Greg S. and Matias,  Yossi and Azizi,  Shekoofeh and Karthikesalingam,  Alan and Natarajan,  Vivek},
  year = {2025},
  pages = {943–950}
}

@article{AMIE-diagnosis,
  title = {Towards accurate differential diagnosis with large language models},
  volume = {642},
  number = {8067},
  journal = {Nature},
  author = {McDuff,  Daniel and Schaekermann,  Mike and Tu,  Tao and Palepu,  Anil and Wang,  Amy and Garrison,  Jake and Singhal,  Karan and Sharma,  Yash and Azizi,  Shekoofeh and Kulkarni,  Kavita and Hou,  Le and Cheng,  Yong and Liu,  Yun and Mahdavi,  S. Sara and Prakash,  Sushant and Pathak,  Anupam and Semturs,  Christopher and Patel,  Shwetak and Webster,  Dale R. and Dominowska,  Ewa and Gottweis,  Juraj and Barral,  Joelle and Chou,  Katherine and Corrado,  Greg S. and Matias,  Yossi and Sunshine,  Jake and Karthikesalingam,  Alan and Natarajan,  Vivek},
  year = {2025},
  pages = {451–457}
}

@article{AMIE-conversation,
  title = {Towards conversational diagnostic artificial intelligence},
  volume = {642},
  number = {8067},
  journal = {Nature},
  author = {Tu,  Tao and Schaekermann,  Mike and Palepu,  Anil and Saab,  Khaled and Freyberg,  Jan and Tanno,  Ryutaro and Wang,  Amy and Li,  Brenna and Amin,  Mohamed and Cheng,  Yong and Vedadi,  Elahe and Tomasev,  Nenad and Azizi,  Shekoofeh and Singhal,  Karan and Hou,  Le and Webson,  Albert and Kulkarni,  Kavita and Mahdavi,  S. Sara and Semturs,  Christopher and Gottweis,  Juraj and Barral,  Joelle and Chou,  Katherine and Corrado,  Greg S. and Matias,  Yossi and Karthikesalingam,  Alan and Natarajan,  Vivek},
  year = {2025},
  month = apr,
  pages = {442–450}
}

@article{MedFound,
  title={A generalist medical language model for disease diagnosis assistance},
  author = {Liu,  Xiaohong and Liu,  Hao and Yang,  Guoxing and Jiang,  Zeyu and Cui,  Shuguang and Zhang,  Zhaoze and Wang,  Huan and Tao,  Liyuan and Sun,  Yongchang and Song,  Zhu and Hong,  Tianpei and Yang,  Jin and Gao,  Tianrun and Zhang,  Jiangjiang and Li,  Xiaohu and Zhang,  Jing and Sang,  Ye and Yang,  Zhao and Xue,  Kanmin and Wu,  Song and Zhang,  Ping and Yang,  Jian and Song,  Chunli and Wang,  Guangyu},
  journal={Nature Medicine},
  volume={31},
  number={3},
  pages={932--942},
  year={2025}
}

@article{NMed-LLMSummary,
  title = {Adapted large language models can outperform medical experts in clinical text summarization},
  volume = {30},
  number = {4},
  journal = {Nature Medicine},
  author = {Van Veen,  Dave and Van Uden,  Cara and Blankemeier,  Louis and Delbrouck,  Jean-Benoit and Aali,  Asad and Bluethgen,  Christian and Pareek,  Anuj and Polacin,  Malgorzata and Reis,  Eduardo Pontes and Seehofnerová,  Anna and Rohatgi,  Nidhi and Hosamani,  Poonam and Collins,  William and Ahuja,  Neera and Langlotz,  Curtis P. and Hom,  Jason and Gatidis,  Sergios and Pauly,  John and Chaudhari,  Akshay S.},
  year = {2024},
  pages = {1134–1142}
}

@article{eval-deepseek,
  title = {Comparative benchmarking of the DeepSeek large language model on medical tasks and clinical reasoning},
  volume = {31},
  number = {8},
  journal = {Nature Medicine},
  author = {Tordjman,  Mickael and Liu,  Zelong and Yuce,  Murat and Fauveau,  Valentin and Mei,  Yunhao and Hadjadj,  Jerome and Bolger,  Ian and Almansour,  Haidara and Horst,  Carolyn and Parihar,  Ashwin Singh and Geahchan,  Amine and Meribout,  Anis and Yatim,  Nader and Ng,  Nicole and Robson,  Phillip and Zhou,  Alexander and Lewis,  Sara and Huang,  Mingqian and Deyer,  Timothy and Taouli,  Bachir and Lee,  Hao-Chih and Fayad,  Zahi A. and Mei,  Xueyan},
  year = {2025},
  pages = {2550–2555}
}

@article{eval-deepseek125Patients,
  title = {Benchmark evaluation of DeepSeek large language models in clinical decision-making},
  volume = {31},
  number = {8},
  journal = {Nature Medicine},
  author = {Sandmann,  Sarah and Hegselmann,  Stefan and Fujarski,  Michael and Bickmann,  Lucas and Wild,  Benjamin and Eils,  Roland and Varghese,  Julian},
  year = {2025},
  pages = {2546–2549}
}

@article{garg2025rise,
  title={The Rise of Small Language Models in Healthcare: A Comprehensive Survey},
  author={Garg, Muskan and Raza, Shaina and Sohn, Sunghwan},
  journal={arXiv preprint arXiv:2504.17119},
  year={2025}
}

@article{builtjes2025leveraging,
  title={Leveraging open-source large language models for clinical information extraction in resource-constrained settings},
  author={Builtjes, Luc and Bosma, Joeran and Prokop, Mathias and van Ginneken, Bram and Hering, Alessa},
  journal={JAMIA Open},
  volume={8},
  number={5},
  pages={ooaf109},
  year={2025},
  doi={10.1093/jamiaopen/ooaf109}
}

@article{zhu2025empowering,
  title={Empowering digital health management with on-device large language models for glucose prediction},
  author={Zhu, Taiyu and Howson, Joanna M. M. and Nevado-Holgado, Alejo},
  journal={medRxiv},
  year={2025},
  doi={10.1101/2025.07.12.25331188}
}

@article{evangelista2026graphrag,
  title={GraphRAG-Enabled Local Large Language Model for Gestational Diabetes Mellitus: Development of a Proof-of-Concept},
  author={Evangelista, Edmund and Ruba, Fathima and Bukhari, Salman and Nazir, Amril and Sharma, Ravishankar},
  journal={JMIR Diabetes},
  volume={11},
  pages={e76454},
  year={2026},
  doi={10.2196/76454}
}

@article{kim2025privacy,
  title={Privacy-by-Design Framework for Large Language Model Chatbots in Urology},
  author={Kim, Eun Joung and Kim, JungYoon},
  journal={International Neurourology Journal},
  volume={29},
  number={Suppl 2},
  pages={S65--S72},
  year={2025},
  doi={10.5213/inj.2550274.137}
}

@article{wang2025jarvis,
  title={Medicine's J.A.R.V.I.S. moment: how DeepSeek-R1 transforms clinical practice},
  author={Wang, Rui and He, Jian and Liang, Heng},
  journal={Journal of Thoracic Disease},
  year={2025},
  doi={10.21037/jtd-25-98153}
}

@article{moell2025deepseek,
  title={Medical Reasoning in LLMs: An In-Depth Analysis of DeepSeek R1},
  author={Moell, Bertil and Aronsson, Filip and Akbar, Shadi},
  journal={arXiv preprint arXiv:2504.00016},
  year={2025}
}

@article{deepseek2025cme,
  title={Evaluating the Performance of DeepSeek-R1 and DeepSeek-V3 Versus OpenAI Models in the Chinese National Medical Licensing Examination: Cross-Sectional Comparative Study},
  author={Wang, Wei and Zhou, Yang and Fu, Jian and Hu, Ke},
  journal={JMIR Medical Education},
  volume={11},
  pages={e73469},
  year={2025},
  doi={10.2196/73469}
}

@article{kocaman2025clever,
  title={Clinical Large Language Model Evaluation by Expert Review (CLEVER): Framework Development and Validation},
  author={Kocaman, Veysel and Kaya, Mustafa Aytu{\u{g}} and Feier, Andrei Marian and Talby, David},
  journal={JMIR AI},
  volume={4},
  pages={e72153},
  year={2025},
  doi={10.2196/72153}
}

@article{croxford2025evaluating,
  title={Evaluating clinical AI summaries with large language models as judges},
  author={Croxford, Ella and Gao, Yanjun and First, Eric and Pellegrino, Nicole and Schnier, Marc and Caskey, James and Oguss, Michael and Wills, Garrett and Chen, Guoyin and Dligach, Dmitriy and Churpek, Matthew and Mayampurath, Anoop and Liao, Fuhai and Goswami, Chaitanya and Wong, Kenrick and Patterson, Blake and Afshar, Majid},
  journal={npj Digital Medicine},
  volume={8},
  number={1},
  pages={1--10},
  year={2025},
  doi={10.1038/s41746-025-02005-2}
}

@article{humaneval2025llmjudge,
  title={Human Evaluators vs. LLM-as-a-Judge: Toward Scalable, Real-Time Evaluation of GenAI in Global Health},
  author={Williams, G. and Rutunda, S. and Nzabakira, F. and Mateen, B.},
  journal={medRxiv},
  year={2025},
  doi={10.1101/2025.10.27.25338910}
}

@article{genovese2026artificial,
  title={Artificial Authority: The Promise and Perils of LLM Judges in Healthcare},
  author={Genovese, Ariana and Hegstrom, Lars and Prabha, Srinivasagam and Gomez-Cabello, Cesar A. and Haider, Syed Ali and Collaco, Bernardo and Wood, Nadia G. and Forte, Antonio Jorge},
  journal={Bioengineering},
  volume={13},
  number={1},
  pages={108},
  year={2026},
  doi={10.3390/bioengineering13010108}
}

@article{davis2026medslice,
  title={MedSlice: fine-tuned large language models for secure clinical note sectioning},
  author={Davis, Joshua and Sounack, Thomas and Sciacca, Kate and Brain, Jessie M and Durieux, Brigitte N and Agaronnik, Nicole D and Lindvall, Charlotta},
  journal={JAMIA Open},
  volume={9},
  number={1},
  pages={ooaf179},
  year={2026},
  doi={10.1093/jamiaopen/ooaf179}
}

@article{guluzade2025elmtex,
  title={ELMTEX: Fine-Tuning Large Language Models for Structured Clinical Information Extraction. A Case Study on Clinical Reports},
  author={Guluzade, Aynur and Heiba, Naguib and Boukhers, Zeyd and Hamiti, Florim and Polash, Jahid Hasan and Mohamad, Yehya and Velasco, Carlos A.},
  journal={arXiv preprint arXiv:2502.05638},
  year={2025}
}

@article{le2025impact,
  title={The Impact of LoRA Adapters on LLMs for Clinical Text Classification Under Computational and Data Constraints},
  author={Le, T.-D. and others},
  journal={IEEE Access},
  volume={13},
  pages={109366--109376},
  year={2025},
  doi={10.1109/ACCESS.2025.3582037}
}

@article{clusmann2025implementing,
  author    = {Dennst{\"a}dt, Fabio and Hastings, Janna and Putora, Paul Martin and Schmerder, Max and Cihoric, Nikola},
  title     = {Implementing large language models in healthcare while balancing control, collaboration, costs and security},
  journal   = {npj Digital Medicine},
  volume    = {8},
  number    = {1},
  pages     = {143},
  year      = {2025},
  doi       = {10.1038/s41746-025-01476-7},
  pmid      = {40050366},
  pmcid     = {PMC11885444}
}

@article{kottlors2025eurorad,
  author    = {Kim, Su Hwan and Schramm, Severin and Adams, Lisa C and Braren, Rickmer and Bressem, Keno K and Keicher, Matthias and Platzek, Paul-S{\"o}ren and Paprottka, Karolin Johanna and Zimmer, Claus and Hedderich, Dennis M and Wiestler, Benedikt},
  title     = {Benchmarking the diagnostic performance of open source {LLMs} in 1933 {Eurorad} case reports},
  journal   = {npj Digital Medicine},
  volume    = {8},
  number    = {1},
  pages     = {97},
  year      = {2025},
  doi       = {10.1038/s41746-025-01488-3},
  pmid      = {39934372},
  pmcid     = {PMC11814077}
}

@article{hager2024limitations,
  author    = {Hager, Paul and Jungmann, Friederike and Holland, Robbie and Bhagat, Kunal and Hubrecht, Inga and Knauer, Manuel and Vielhauer, Jakob and Makowski, Marcus and Braren, Rickmer and Kaissis, Georgios and Rueckert, Daniel},
  title     = {Evaluation and mitigation of the limitations of large language models in clinical decision-making},
  journal   = {Nature Medicine},
  volume    = {30},
  number    = {9},
  pages     = {2613--2622},
  year      = {2024},
  doi       = {10.1038/s41591-024-03097-1},
  pmid      = {38965432},
  pmcid     = {PMC11405275}
}

@article{qiu2025medrench,
  author    = {Qiu, Pengcheng and Wu, Chaoyi and Liu, Shuyu and Fan, Yanjie and Zhao, Weike and Chen, Zhuoxia and Gu, Hongfei and Peng, Chuanjin and Zhang, Ya and Wang, Yanfeng and Xie, Weidi},
  title     = {Quantifying the reasoning abilities of {LLMs} on clinical cases},
  journal   = {Nature Communications},
  volume    = {16},
  number    = {1},
  pages     = {9799},
  year      = {2025},
  doi       = {10.1038/s41467-025-64769-1},
  pmid      = {41198657},
  pmcid     = {PMC12592457}
}

@article{thirunavukarasu2023llmedicine,
  author    = {Thirunavukarasu, Arun James and Ting, Darren Shu Jeng and Elangovan, Kabilan and Gutierrez, Laura and Tan, Ting Fang and Ting, Daniel Shu Wei},
  title     = {Large language models in medicine},
  journal   = {Nature Medicine},
  volume    = {29},
  number    = {8},
  pages     = {1930--1940},
  year      = {2023},
  doi       = {10.1038/s41591-023-02448-8},
  pmid      = {37460753}
}

@article{zhang2025ophthreview,
  author    = {Zhang, Zili and Zhang, Haiyang and Pan, Zhe and Bi, Zhangqian and Wan, Yao and Song, Xuefei and Fan, Xianqun},
  title     = {Evaluating Large Language Models in Ophthalmology: Systematic Review},
  journal   = {Journal of Medical Internet Research},
  volume    = {27},
  pages     = {e76947},
  year      = {2025},
  doi       = {10.2196/76947},
  pmid      = {41144954},
  pmcid     = {PMC12603593}
}

@article{goh2025gpt4rct,
  author    = {Goh, Ethan and Gallo, Robert J and Strong, Eric and Weng, Yingjie and Kerman, Hannah and Freed, Jason A and Cool, Jos{\'e}phine A and Kanjee, Zahir and Lane, Kathleen P and Parsons, Andrew S and Ahuja, Neera and Horvitz, Eric and Yang, Daniel and Milstein, Arnold and Olson, Andrew P J and Hom, Jason and Chen, Jonathan H and Rodman, Adam},
  title     = {{GPT-4} assistance for improvement of physician performance on patient care tasks: a randomized controlled trial},
  journal   = {Nature Medicine},
  volume    = {31},
  number    = {4},
  pages     = {1233--1238},
  year      = {2025},
  doi       = {10.1038/s41591-024-03456-y},
  pmid      = {39910272},
  pmcid     = {PMC12380382}
}

@article{gao2025loraperiop,
  author    = {Gao, Shaowei and Zhao, Xu and Chen, Lihui and Yu, Junrong and Tian, Shuning and Zhou, Huaqiang and Chen, Jingru and Long, Sizhe and He, Qiulan and Feng, Xia},
  title     = {Enhancing privacy-preserving deployable large language models for perioperative complication detection: a targeted strategy with {LoRA} fine-tuning},
  journal   = {npj Digital Medicine},
  volume    = {8},
  number    = {1},
  pages     = {773},
  year      = {2025},
  doi       = {10.1038/s41746-025-02139-3},
  pmid      = {41390570},
  pmcid     = {PMC12717251}
}

@inproceedings{hu2022lora,
  title     = {{LoRA}: Low-Rank Adaptation of Large Language Models},
  author    = {Hu, Edward J. and Shen, Yelong and Wallis, Phillip and Allen-Zhu, Zeyuan and Li, Yuanzhi and Wang, Shean and Wang, Lu and Chen, Weizhu},
  booktitle = {International Conference on Learning Representations},
  year      = {2022},
  url       = {https://openreview.net/forum?id=nZeVKeeFYf9}
}

@article{bhayana2024radiology,
  author    = {Bhayana, Rajesh},
  title     = {Chatbots and Large Language Models in Radiology: {A} Practical Primer for Clinical and Research Applications},
  journal   = {Radiology},
  volume    = {310},
  number    = {1},
  pages     = {e232756},
  year      = {2024},
  doi       = {10.1148/radiol.232756},
  pmid      = {38226883}
}

@article{schramm2025multimodal,
  author    = {Schramm, Severin and Preis, Silas and Metz, Marie-Christin and Jung, Kirsten and Schmitz-Koep, Benita and Zimmer, Claus and Wiestler, Benedikt and Hedderich, Dennis M and Kim, Su Hwan},
  title     = {Impact of Multimodal Prompt Elements on Diagnostic Performance of {GPT-4V} in Challenging Brain {MRI} Cases},
  journal   = {Radiology},
  volume    = {314},
  number    = {1},
  pages     = {e240689},
  year      = {2025},
  doi       = {10.1148/radiol.240689},
  pmid      = {39835982}
}

@article{asgari2025hallucination,
  author    = {Asgari, Elham and Monta{\~n}a-Brown, Nina and Dubois, Magda and Khalil, Saleh and Balloch, Jasmine and Yeung, Joshua Au and Pimenta, Dominic},
  title     = {A framework to assess clinical safety and hallucination rates of {LLMs} for medical text summarisation},
  journal   = {npj Digital Medicine},
  volume    = {8},
  number    = {1},
  pages     = {274},
  year      = {2025},
  doi       = {10.1038/s41746-025-01670-7},
  pmid      = {40360677},
  pmcid     = {PMC12075489}
}

@article{omar2025adversarial,
  author    = {Omar, Mahmud and Sorin, Vera and Collins, Jeremy D and Reich, David and Freeman, Robert and Gavin, Nicholas and Charney, Alexander and Stump, Lisa and Bragazzi, Nicola Luigi and Nadkarni, Girish N and Klang, Eyal},
  title     = {Multi-model assurance analysis showing large language models are highly vulnerable to adversarial hallucination attacks during clinical decision support},
  journal   = {Communications Medicine},
  volume    = {5},
  number    = {1},
  pages     = {330},
  year      = {2025},
  doi       = {10.1038/s43856-025-01021-3},
  pmid      = {40753316},
  pmcid     = {PMC12318031}
}

@article{omiye2024pitfalls,
  author    = {Omiye, Jesutofunmi A and Gui, Haiwen and Rezaei, Shawheen J and Zou, James and Daneshjou, Roxana},
  title     = {Large Language Models in Medicine: {T}he Potentials and Pitfalls: {A} Narrative Review},
  journal   = {Annals of Internal Medicine},
  volume    = {177},
  number    = {2},
  pages     = {210--220},
  year      = {2024},
  doi       = {10.7326/M23-2772},
  pmid      = {38285984}
}

@article{wind2025rar,
  author    = {Wind, Sebastian and Sopa, Jeta and Truhn, Daniel and Lotfinia, Mahshad and Nguyen, Tri-Thien and Bressem, Keno and Adams, Lisa and Rusu, Mirabela and K{\"o}stler, Harald and Wellein, Gerhard and Maier, Andreas and Tayebi Arasteh, Soroosh},
  title     = {Multi-step retrieval and reasoning improves radiology question answering with large language models},
  journal   = {npj Digital Medicine},
  volume    = {8},
  number    = {1},
  pages     = {790},
  year      = {2025},
  doi       = {10.1038/s41746-025-02250-5},
  pmid      = {41429891},
  pmcid     = {PMC12749912}
}

@article{chen2021codex,
  title   = {Evaluating Large Language Models Trained on Code},
  author  = {Chen, Mark and Tworek, Jerry and Jun, Heewoo and Yuan, Qiming and Pinto, Henrique Ponde de Oliveira and Kaplan, Jared and Edwards, Harri and Burda, Yuri and Joseph, Nicholas and Brockman, Greg and Ray, Alex and Puri, Raul and Krueger, Gretchen and Petrov, Michael and Khlaaf, Heidy and Sastry, Girish and Mishkin, Pamela and Chan, Brooke and Gray, Scott and Ryder, Nick and Pavlov, Mikhail and Power, Alethea and Kaiser, Lukasz and Bavarian, Mohammad and Winter, Clemens and Tillet, Philippe and Such, Felipe Petroski and Cummings, Dave and Plappert, Matthias and Chantzis, Fotios and Barnes, Elizabeth and Herbert-Voss, Ariel and Guss, William Hebgen and Nichol, Alex and Paino, Alex and Tezak, Nikolas and Tang, Jie and Babuschkin, Igor and Balaji, Suchir and Jain, Shantanu and Saunders, William and Hesse, Christopher and Carr, Andrew N. and Leike, Jan and Achiam, Josh and Misra, Vedant and Morikawa, Evan and Radford, Alec and Knight, Matthew and Brundage, Miles and Murati, Mira and Mayer, Katie and Welinder, Peter and McGrew, Bob and Amodei, Dario and McCandlish, Sam and Sutskever, Ilya and Zaremba, Wojciech},
  journal = {arXiv preprint arXiv:2107.03374},
  year    = {2021}
}

\end{document}